\RequirePackage{fix-cm}
\documentclass[twocolumn]{svjour3}          % twocolumn
\smartqed  % flush right qed marks, e.g. at end of proof
\usepackage{newtxtext}		% use Times fonts if available on your TeX system
\usepackage{graphicx}
\usepackage{marvosym}
\usepackage{amssymb}
\usepackage{amsmath}
\usepackage{mathrsfs}
\usepackage{tabu}
\usepackage{tabularx}
\usepackage{booktabs}
\usepackage[linesnumbered,lined,boxed,commentsnumbered]{algorithm2e}
\usepackage{todonotes}
\usepackage{bbm}
\usepackage{multirow}
\usepackage{hyperref}
\usepackage{listings}
\usepackage{color}
\usepackage{scrextend}
\usepackage{tikz}

\usepackage{cite}

\DeclareMathOperator*{\argmin}{arg\,min}

\begin{document}

%\title{Efficient and Accurate Discovery of Dynamic Condition Response Graphs}
% If the results for Minerful are true, I propose the title below, otherwise the title above.
\title{DisCoveR: Accurate \& Efficient Discovery of Declarative Process Models}

%\title{Efficient Discovery of Dynamic Condition Response Relations \\ (other title suggestions?)}%\thanks{Grants or other notes

%\subtitle{Mining Dynamic Condition Response Graphs with DisCoveR}


\author{Christoffer Olling Back$^\text{1}$\and
Tijs Slaats$^\text{1}$\and
Thomas Troels Hildebrandt$^\text{1}$\and
Morten Marquard$^\text{2}$\thanks{Work supported by the Innovation Fund Denmark project \emph{EcoKnow} (7050-00034A) and the Danish Council for Independent Research project \emph{Hybrid Business Process Management Technologies} (DFF-6111-00337)}
}

\authorrunning{C. O. Back, T. Slaats, T. T. Hildebrandt, M. Marquard} % if too long for running head

\institute{ 
    \begin{tabular}{l l}
        \Letter &  Christoffer Olling Back \\
        & back@di.ku.dk \\
        & \includegraphics[width=2.5mm]{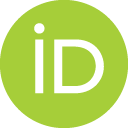}~https://orcid.org/0000-0001-7998-7167 \\
        \\
        & Tijs Slaats \\
        & slaats@di.ku.dk \\
        & \includegraphics[width=2.5mm]{orcid.png}~https://orcid.org/0000-0002-7435-5563 \\
        \\
        & Thomas Troels Hildebrandt \\
        & hilde@di.ku.dk \\
        & \includegraphics[width=2.5mm]{orcid.png}~https://orcid.org/0000-0001-6244-6970 \\
        \\
        & Morten Marquard \\
        & mm@dcrsolutions.net \\
        \\
        \\
        $^\text{1}$ & Department of Computer Science, University of Copenhagen \\ 
        & Copenhagen, Denmark \\
        $^\text{2}$ & DCR Solutions, Copenhagen, Denmark
    \end{tabular}
}

\date{Received: date / Accepted: date}

\maketitle

\begin{abstract}
\emph{Declarative} process modeling formalisms - which capture high-level process constraints - have seen growing interest, especially for modeling flexible processes. This paper presents DisCoveR, an extremely efficient and accurate declarative miner for learning Dynamic Condition Response (DCR) Graphs from event logs. We precisely formalize the algorithm, describe a highly efficient bit vector implementation and rigorously evaluate performance against two other declarative miners, representing the state-of-the-art in Declare and DCR Graphs mining. DisCoveR outperforms each of these w.r.t. a binary classification task, achieving an average accuracy of 96.2\% in the Process Discovery Contest 2019. Due to its linear time complexity, DisCoveR also achieves run-times 1-2 \emph{orders of magnitude} below its declarative counterparts.

Finally, we show how the miner has been integrated in a state-of-the-art declarative process modeling framework as a model recommendation tool, discuss how discovery can play an integral part of the modeling task and report on how the integration has improved the modeling experience of end-users.

\end{abstract}

\newcommand{\condition}{\rightarrow\hspace{-3pt}\bullet}
\newcommand{\response}{\bullet\hspace{-3pt}\rightarrow}
\newcommand{\legal}{$+$}
\newcommand{\illegal}{$-$}
\newcommand{\minerful}{MINERful}
\newcommand{\baselineminer}{Debois, et al's miner}
\newcommand{\discover}{DisCoveR}
\newcommand{\atmostone}{\textsc{AtMostOne}}
\newcommand{\responsedeclare}{\textsc{Response}}
\newcommand{\precedence}{\textsc{Precedence}}
\newcommand{\alternateprecedence}{\textsc{AlternatePrecedence}}
\newcommand{\chainprecedence}{\textsc{ChainPrecedence}}
\newcommand{\notchainsuccession}{\textsc{NotChainSuccession}}
\newcommand{\notcoexistence}{\textsc{NotCoExistence}}
\newcommand{\succession}{\textsc{Succession}}
\newcommand{\chainsuccession}{\textsc{ChainSuccession}}

\section{Introduction}
\label{sec:introduction}
Technologies for business process management have matured significantly since the early proposals of office automation systems and business process definition languages in the late 1970s~\cite{hajobook,DBLP:books/mit/Aalst2002,DBLP:books/daglib/0029914}. Today, BPMN~\cite{HagenBPMNuse,bpmn2.0normative2011,dijkman2008semantics} has become a stable, de-facto standard notation for describing business processes. 
Users can choose from a number of commercial design tools and business process management systems, supporting the design and enactment of business processes. In the recent years, we have even seen commercial process mining tools~\cite{la2011apromore}, supporting the automated discovery of BPMN models from event logs that record traces from historical processes~\cite{van2004workflow,van2011process}. 

However, the imperative approach to business process management, which has dominated the development of business process management technologies for the last 50 years, falls short when it comes to automating or supporting knowledge intensive processes~\cite{DiCiccio2015,SantosFranca2015} that need flexibility in executions, but still need to adhere to rules~\cite{pesic2007declare,mukkamala2012formal,tijsthesis}. 

Governmental case work processes are particular challenging examples of such constrained knowledge work processes, since the development of new laws and changes to existing laws gives rise to changes in the rules, and most often an increase in their complexity~\cite{6975351,6037570}. This makes it difficult to define and maintain standardized processes using the imperative approach: since commercial tools do not support verification of compliance with the law, the decision whether a standardized process is still compliant with the law and if not, how to make it compliant, becomes a manual and error-prone task. Also, if the case is critical in nature, as in healthcare or social services, citizens would expect their cases to follow an individualized path that helps them best, not a standardized process dictated by what is deemed possible by the technology in use.

To combat this challenge, it has been proposed to use declarative notations~\cite{DBLP:conf/edoc/PesicSA07,DBLP:journals/corr/abs-1110-4161,10.1007/978-3-319-23063-4_15} for formalizing the rules governing knowledge work processes. Indeed, the use of declarative models can be seen as an alternative line of work in the area of computer supported work processes that can be traced back to the early rule-based expert systems~\cite{giarratano1998expert}. A key challenge of the declarative approach has since then been the lack of a standard, understandable formal notation, that could be used by domain experts to formalize the rules to be followed and enacted by rule engines supporting the case worker. 

This challenge was taken up in the EcoKnow (Effective, co-created and compliant adaptive case management for Knowledge workers) inter-disciplinary research project initiated in September 2017 and running for four years.

The scientific basis of the project spans from field studies of case management and computer supported cooperative work, over development of workflow technologies based on formal declarative models to machine learning and studies of the understandability of modeling notations. The aim is to provide technologies for flexible, agile and transparent digitalisation of governmental case management processes that at the same time increase quality with respect to compliance and equal treatment of citizens with respect to the law while reducing wasteful delays in case management. 

The technological foundation of the project is the formal Dynamic Condition Response (DCR) Graphs notation~\cite{hildebrandt2011declarative,mukkamala2012formal,tijsthesis}. DCR Graphs is a declarative notation developed during the last decade with the aim of providing flexible process support. The notation is supported by a commercial design tool~\cite{10.1007/978-3-319-23063-4_15} and a stand-alone rule engine, which has can be embedded in third-party case management tools. In particular, the engine has successfully been embedded in two commercial case management solutions developed by the company KMD, which is part of the multi-national company NEC. One of the solutions is used in Danish municipalities, e.g. for handling building permits, the other is used in more than  70\% of Danish central government institutions, including law enforcement, military, tax authorities and largest public universities\footnote{http://www.kmd.dk/indsigter/fleksibilitet-og-dynamisk-sagsbehandling-i-staten.}.

Currently, the engine is being embedded in a case management system delivered by Fujitsu to more than 10\% of Danish municipalities and is also supported by an open source case management tool, Open Case Manager, that can be integrated with third-party document management systems. The Open Case Manager is used both for research demonstration purposes and for actual case management in one of the municipalities participating in the EcoKnow project. 

A key hypothesis of the project is that the declarative DCR Graph modeling notation supports an agile, incremental digitalisation of workflows that, if supported by the right tools, can be understood, developed and maintained locally in governmental institutions instead of relying on vendors of case management solutions to encode laws, and other rules to be followed, into their solutions. Being now half way through the project, the hypothesis has been partially validated. 
%The DCR technology has successfully been embedded in %commercial case management systems used widely in %both local and central government institutions in %Denmark. The technology has also made available as %part of an open source case management tool, the Open %Case Manager, which is now used in the daily case %management in a Danish municipality taking part of %the project. 

To support the local development and maintenance of the declarative DCR models, several modeling tools have been developed~\cite{debois_dcr_2017,10.1007/978-3-319-23063-4_15,8536151}, supported by formal understandability studies~\cite{10.1007/978-3-030-11641-5_37,10.1007/978-3-030-20618-5_5,10.1007/978-3-030-33223-5_14}. 
%A key such tool is the process highlighter~\cite{}, which allows domain experts to define DCR models by highlighting activities, roles and constraints in textual representations of regulations, e.g. law texts, thereby providing traceability between textual representations of rules, e.g. law texts, and DCR models. 
Along with the tools, a methodology for modeling with DCR has been developed, advocating an iterative and incremental, scenario-driven approach with three main tasks. First, to identify key activities and roles. Second, to perform simulations of wanted and unwanted scenarios. Finally, the modeler may either go back to add missing activities and roles or forward to the task of identifying rules that supports the wanted scenarios and forbid the unwanted scenarios.
%In this way, the rules can be digitalized incrementally while the system is in use, instead of up front digitalising an entire law consisting of hundreds or perhaps even thousands of pages.

The iterative approach lends itself extremely well to being supported by process discovery: after the users define wanted and unwanted scenarios, discovery algorithms can be used to automatically make suggestions for which rules should be added. Such a discovery algorithm needs to be both efficient and accurate. On the one hand users expect their modeling experience to be continuous without long interruptions waiting for a discovery algorithm to compute possible rules. On the other hand, they are only helped by rule suggestions that are relevant and correct in terms of the suggested scenarios: poor suggestions will only confuse the users and reduce the quality of their modeling experience.

Recently such an efficient and accurate discovery algorithm was developed for DCR Graphs and afterwards implemented in the commercial design tool~\cite{caisedcrmining}. A remarkable feature of the algorithm is that it can provide accurate and useful suggestions of constraints even with a very few traces as a training set. This means that it can not only be used to discover rules from large logs of historical traces as traditional process mining algorithms, but also be used for recommending constraints based on a few simulated scenarios carried out as part of the scenario-driven modeling approach. In addition the algorithm runs in linear time (with respect to the number of events in the log) with a surprisingly small constant factor, providing near-instantaneous feedback based on user scenarios. 

This paper is part of a special issue of the journal in connection with the Process Discovery Contest 2019, where the high accuracy of the algorithm managed to secure it a second place.
The algorithm itself was first introduced by Nekrasaite et al. in ~\cite{caisedcrmining}, the current paper expands on this initial  introduction with: (1) a complete and thorough formalization of the algorithm that provides all details required for its implementation (Section~\ref{sec:algorithm}); (2) a novel, open source and more efficient implementation based on bit vector operations (Section~\ref{sec:implementation}); (3) a novel evaluation of the algorithm based on the classification task provided by the Process Discovery Contest 2019, showing that the algorithm is currently \emph{the} front-runner in terms of accuracy in declarative process discovery  (Section~\ref{sec:evaluation}); (4) an evaluation of the efficiency of the novel implementation, showing that it is one order of magnitude more efficient than the state-of-the-art in DCR Graphs mining and two orders of magnitude more efficient than the state-of-the-art in Declare mining (Section~\ref{sec:evaluation}); (5) a case study showing how the algorithm has been swiftly transferred to industry through its integration in the dcrgraphs.net process modeling portal, leading to an enhanced modeling experience by its users (Section~\ref{sec:casestudies}).

We proceed as sketched above and in addition we discuss related work in Section~\ref{sec:relatedwork}, preliminaries in Section~\ref{sec:preliminaries}, and conclude and propose future directions of research in Section~\ref{sec:conclusion}.

%
%This can be seen in the reoccurring, unsuccessful %attempts of simplification of the law, such as the %recent agreement by all political parties in Denmark %that existing laws should be simplified where %possible and all new laws should be made %digitalisation ready.

\section{Related Work}
\label{sec:relatedwork}

Many declarative process notations have been developed, several with corresponding discovery algorithms~\cite{jodsjournaltijs}. 
The first of these was Declare~\cite{declare,DecSerFlow:2006,DBLP:conf/edoc/PesicSA07}, which was inspired by property specification patterns for linear temporal logic (LTL)\cite{dwyer1999patterns}.
Declare identified a particular set of patterns relevant for business processes and gave them semantics through a mapping to LTL formulae relevant for describing the rules governing a business process. A Declare model is therefore a collection of such patterns, and the semantics of a model is defined as the traces that satisfy the conjunction of the formulae underlying the patterns.
More recently the same patterns have been formalized using colored automata~\cite{DECLAREBPM2011}, SCIFF~\cite{montali2010declarative,montali331}, and regular expressions \cite{Westergaard2013UnconstrainedMinerED}.
Extensions to Declare include timed~\cite{dectime} and data~\cite{decdata} constraints, which were combined in MP-Declare~\cite{mpdec} (Multi Perspective Declare), and hierarchy~\cite{Zugal2015}.
The first miner for Declare was the Declare Maps Miner~\cite{5949297}, while initially using a brute-force approach, it was extended with several improvements~\cite{10.1007/978-3-642-31095-9_18} inspired by the Apriori algorithm for association rule mining~\cite{Agrawal:1994:FAM:645920.672836}. More recently the miner was extended to allow for parallelization~\cite{MAGGI2018136}.
The second Declare miner to be developed was Minerful~\cite{Ciccio:2015:DDC:2677016.2629447} which provided significant gains in efficiency. Since its introduction it has been extended with support for target-branched constraints~\cite{DiCiccio:2016:EDT:2869182.2869507}, removal of redundancies and inconsistencies~\cite{DICICCIO2017425} and removal of vacuously satisfied constraints~\cite{DICICCIO2018144}.

Another prominent declarative approach is the Guard-Stage-Milestone (GSM) notation~\cite{Hull:2010:IGA:1987781.1987782}, inspired by earlier work on artifact-centric business processes~\cite{Bhattacharya07towardsformal}. GSM aims to effectively model case management and has been a primary contributor to the development of the Case Management Model And Notation (CMMN)~\cite{CMMN}. 
CMMN has seen a relatively fast industrial and academic adoption through the development of tools and case studies~\cite{kurz2015leveraging,wiemuth2017application,herzberg2014modeling}.
Work on process discovery for GSM or CMMN on the other hand is still rather sparse, only one discovery algorithm has been proposed to date~\cite{doi:10.1142/S021884301550001X} with no working implementation. 

Process discovery has also been considered for the Declarative Process Intermediate Language (DPIL)~\cite{schonig2015dpil,zeising2014towards}, which is a  textual, multi-perspective, declarative modeling language. 
Process discovery for DPIL is supported through the DPIL Miner\footnote{http://www.kppq.de/miner.html}. In comparison to other Declarative miners, which tend to focus on the control-flow perspective of processes, the DPIL Miner instead focuses more on mining the organizational perspective~\cite{schonig2016framework}. Interestingly the miner has never been made publicly available and its effectiveness or accuracy can not be independently ascertained.

In more recent work it has been proposed to combine declarative and imperative discovery to produce so-called hybrid~\cite{pocketsflex,10.1007/978-3-319-48472-3_32,Debois2018} or mixed~\cite{10.1007/978-3-642-40176-3_24,10.1007/978-3-319-19069-3_6,DeSmedt2016} models that combine both paradigms. Hybrid miners include the Fusion miner~\cite{DESMEDT2015123}, which produces an inter-mixed Petri net and Declare model, the Hybrid Miner~\cite{10.1007/978-3-319-10172-9_27}  which produces a hierarchical Petri net and Declare model, and the Precision Optimization Hybrid Miner~\cite{10.1007/978-3-319-93931-5_14} which produces a process tree in which some nodes may be Declare models.

%The most well-known declarative workflow modelling formalism is Declare. First presented in \cite{pesic2007declare}, Declare consist of a set of 20 Linear Temporal Logic (LTL) templates. Declare has since been extended to include timing constraints using Metric Interval Temporal Logic and timed automata \cite{westergaard2012looking}, as well as quantifying over data using Metric First Order Temporal Logic in \cite{burattin2016conformance}. The first miner for Declare, the Declare Maps Miner, was presented in \cite{maggi2013declarative}. \minerful~, which also discovers Declare-based models, makes a number of efficiency improvements \cite{ciccio2015discovery}. 

Approaches to workflow formalization based on Classical Linear Logic, a resource-aware logic, were implemented in WorkFlowFM \cite{papapanagiotou2017workflowfm,papapanagiotou2018pragmatic}~ which guarantees concurrent, correct-by-construction processes. The framework was applied to intra-hospital patient transfers in \cite{manataki2016workflow}.

Temporal logics have also been used to model phenomena which would not be considered workflows, such as robot motion \cite{fu2015computational}, naval traffic and train network monitoring \cite{kong2016temporal}.

Process mining is often framed as an inherently \emph{descriptive} rather than \emph{predictive} data mining problem, which precludes the use of standard evaluation metrics familiar in classification and regression tasks. This is largely due to the assumption that an event log represents only positive examples \cite{goedertier2007process}. Some authors have addressed this by developing technique to generate artificial negative examples \cite{goedertier2009robust}.

%Finally, interactive and guided process mining process mining has been adressed in \cite{}. \ldots

Finally, DCR Graphs were inspired by event structures~\cite{Nielsen1979} and developed after Declare was shown to not be sufficiently expressive in modeling industrial cases~\cite{DDBP:2008}. In contrast to Declare, the semantics of DCR Graphs are defined as transformations on the markings of the events. This allows modellers to straightforwardly reason about the execution semantics of a model by simulating it and observing the changes to the markings as events are executed.~\cite{10.1007/978-3-319-23063-4_15} 
Since their inception DCR Graphs have been extended with nesting~\cite{fsen:2011}, time~\cite{flacos2013}, data~\cite{raothesis,10.1007/978-3-642-40176-3_28,8536151}, and hierarchy~\cite{10.1007/978-3-319-10172-9_2}.

%Dynamic Condition Response Graphs \cite{mukkamala2012formal,hildebrandt2011declarative} were extended to included spawning of subprocesses in \cite{debois2015safety}, and formed the basis of a data-oriented declarative language called RESEDA \cite{seco2018reseda}. The only other mining algorithm for DCR Graphs than that presented here was developed by Debois, et. al. in \cite{debois2017declarative}.

%A key such tool is the process highlighter~\cite{}, which allows domain experts to define DCR models by highlighting activities, roles and constraints in textual representations of regulations, e.g. law texts, thereby providing traceability between textual representations of rules, e.g. law texts, and DCR models. 

\section{Preliminaries}
\label{sec:preliminaries}

We briefly recall the formal definitions of processes, event logs, and give a formal presentation of the task of process discovery, as well as the DCR Graphs formalism.

\begin{definition}[Processes and Event Logs]
  \begin{itemize}
  	\item An \emph{alphabet} $\Sigma$ is a finite set of symbols denoting activities. We denote by $\Sigma_L$ activities present in log $L$.
	\item $\Sigma^+$ denotes the countably infinite set of finite, nonempty strings, i.e. sequences, over $\Sigma$.
	\item A \emph{process} is a pair $(P, \mathbb{P}_P)$ where $P$ is a set of allowable sequences of activities, i.e. $P \subseteq \Sigma^+$ along with an associated probability distribution $\mathbb{P}_P$ over $P$
	\item An \emph{event}, denoted $\varsigma$, is a particular occurrence of an activity.
	\item A \emph{trace} $\sigma \in \Sigma^+ = \langle \varsigma_1,\ldots, \varsigma_i, \ldots, \varsigma_n \rangle$ represents a sequence of activities, with $i \in \mathbb{N}$. A trace can be seen as a partial mapping:
	 $$\sigma(i): \mathbb{N} \hookrightarrow \Sigma$$
  	\item A \emph{process model} $h$ defines a semantics such that the \emph{language} $\ell$ of $h$ denotes the set of traces accepted by $h$. That is, 
  	$$\ell(h) \subseteq \Sigma^+$$
	\item Finally, a \emph{log} $L$ is a multiset representing the number of occurrences of different traces: 
		$$ L = \left\{ \sigma_1^{m(\sigma_1)}, \ldots, \sigma_n^{m(\sigma_n)} \right\}$$ 
	where $m(\sigma_k) \in \mathbb{N}$ denotes the multiplicity of $\sigma_k$. A log can be seen as a sample from $(P, \mathbb{P}_P)$.
  \end{itemize} 
\end{definition}

Note the assumption of strict monotonicity implied by this definition of traces. That is, for all $i,j \in \mathbb{N}$ we have that

$$
i < j \implies \sigma(i) \prec \sigma(j) 
$$

where $\prec$ denotes ``precedes'', and also that

$$
i = j \implies \sigma(i) = \sigma(j). 
$$

This means that no distinct events can share the exact same timestamp.

\begin{definition}[Process Discovery]
    \emph{Process discovery} refers to a procedure that derives a process model from an event log. Let $\mathcal{L}$ denote the set of all valid event logs and $\mathcal{H}_F$ the set of process models encodable by some process modeling formalism $F$. A process discovery algorithm $\gamma$ is a mapping: 
    $$
    \gamma: \mathcal{L} \rightarrow \mathcal{H}_F 
	$$
 
Examples of $F$ include Petri nets, sound Petri nets, WorkFlow nets, R/I-nets, Declare maps, and of course DCR Graphs. In other words, $\mathcal{H}_F$ is our \emph{hypothesis space} to which our learning algorithm is restricted.

By extension, we can view the overall task as a mapping from a log to a language, i.e.~a subset of all possible traces:

$$
\ell(\gamma): \mathcal{L} \rightarrow 2^{\Sigma^+} 
$$

Where $2^\mathcal{X}$ denotes the \emph{powerset} of set $\mathcal{X}$. To see this, consider that for some $L \in \mathcal{L}$, we have $\gamma(L) = h$ and $\ell(h) \subseteq 2^{\Sigma^+}$.

\end{definition}

%\begin{definition}[Flower Model] 
%\todo{We might consider removing thisif we don't discuss normalized precision} If $\ell(h) = \Sigma^+$, that is if the process model $h$ allows all possible behavior resulting from the activities in $\Sigma$, then $h$ is called a \emph{flower model}. This definition is clearly universal to all modeling formalisms.
%\end{definition} 

\begin{definition}[DCR Graphs]
\label{def:dcr}
DCR Graphs consist of a set of events with three associated unary predicates: \emph{executed}, \emph{pending}, and \emph{included}. Moreover, four binary relations are defined between events. In order to be executed, an event must be included and satisfy any associated relations.

Formally, a \emph{dynamic condition response graph} is a tuple

$$g = ( \mathcal{E}, m, A, \response, \condition, +, \%, l)$$

where

\begin{itemize}
 	\item $\mathcal{E}$ is a set of events.
	\item $m \in \mathscr{M}(g) = 2^\mathcal{E} \times 2^\mathcal{E} \times 2^\mathcal{E}$ is the \emph{marking} $\mathscr{M}(g)$ is the set of all \emph{markings}.
	\item $A$ is the set of  \emph{activities}.
	\item $\condition \hspace{3pt} \in \mathcal{E} \times \mathcal{E}$ is the \emph{condition} relation.
	\item $\response \hspace{3pt} \in \mathcal{E} \times \mathcal{E}$ is the \emph{response} relation.
	\item $+$ is the \emph{includes} relation
	\item $\%$ is the \emph{excludes} relation
	\item $+ \cap \% = \emptyset$
	\item $l: \mathcal{E} \rightarrow A$ is a labeling function mapping every event to an activity. 
\end{itemize}
\end{definition}

A DCR Graph marking $m = (\mathsf{Ex}, \mathsf{Pe}, \mathsf{In})$ represents events which have previously been \emph{executed}, \emph{pending} events to be executed or excluded, and events currently \emph{included}. For finite traces, a DCR Graph is defined to be \emph{accepting} when $\mathsf{Pe} \cap \mathsf{In} = \emptyset$, i.e.~no pending events are currently included.

The execution semantics of DCR Graphs requires that for an event $e$ to be executed, it must fulfill the following criteria:
\begin{itemize}
  \item $e$ must be \emph{included}, i.e.~$e \in \mathsf{In}$
  \item If any condition relations exist s.t. $e' \condition e$, then all such $e'$ must have been executed, \emph{or excluded}, i.e.~$e' \in \mathsf{Ex}$ or $e' \in \mathsf{In}^C$
\end{itemize}

Furthermore, if $e$ is executed, the marking $m$ will change as follows:

\begin{itemize}
  \item If any response relations exist s.t. $e \condition \hspace{3pt} e'$, then all such $e'$ will become \emph{pending}.
  \item If any excludes relations exist s.t. $e \rightarrow\hspace{-6pt}\% \hspace{3pt} e'$ then any included $e'$ will become \emph{excluded}.
  \item If any includes relations exist s.t. $e \rightarrow\hspace{-6pt}+ \hspace{3pt} e'$ then any excluded $e'$ will become \emph{included}.
\end{itemize}

An important point to note regards the labeling function $l$. As $l$ may map more than one event to the same activity, which can potentially result in a non-deterministic model. In the algorithm presented here, only injective labeling functions are considered, so each event is mapped to exactly one activity.

\section{Algorithm}
\label{sec:algorithm}

In this section we formally describe the ParNek algorithm, implementation details are described in Section \ref{sec:implementation}. The algorithm always produces perfectly fitting models, i.e.~all traces in the log will be replayable on the generated model. The algorithm proceeds in following the following steps:

\begin{enumerate}
  \item A set of candidates for four relation patterns is constructed.
  \item Additional excludes relations are added based on predecessor and successor relations.
  \item Additional includes/excludes patterns are added analogous to a \notchainsuccession~ relations.
  \item Redundant excludes relations are removed.
    \item Redundant condition and response relations are removed via transitive reduction.
  \item Additional condition relations are discovered using a limited replay strategy.
  \item A final transitive reduction is performed for condition relations.
\end{enumerate}

%The algorithm allows for a total of 32 variants w.r.t. the 5 boolean parameters which determine: the strategy for finding inclusions and exclusions, whether to include additional conditions, and whether certain optimizations, and reductions are to be applied.

We will refer to seven relation templates from the LTL-based modeling language \emph{Declare}. The relations are described in words in Table \ref{tab:parnek_core} with analogous DCR relations. Precise formal definitions of functions for identifying relations satisfied by the log are given in Table \ref{tab:functions}. We refer to lines in the high-level control flow pseudocode in Algorithm \ref{ParNek}.

The first step of the ParNek algorithm is the initialization of a DCR Graph, after which we begin adding relations using a number of strategies.

\begin{table*}
\normalsize
\begin{tabu} to \textwidth { X[0.65] X[0.5] X[1.0] }
\toprule
Declare 							& DCR Graphs 										& Description \\
\midrule 
\atmostone$(a)$ 			& $a\rightarrow\hspace{-4pt}\%\hspace{3pt} a$ 		& Activity $a$ can occur $0$ or $1$ time \\
\responsedeclare$(a, b)$ 			& $a \hspace{4pt} \response \hspace{1pt} b$			& After $a$ occurs, $b$ must eventually occur\\
\precedence$(a, b)$ 		& $a \hspace{1pt} \condition \hspace{4pt} b$		& Before $b$ can occur, $a$ must have occurred\\
\alternateprecedence$(a, b)$& $a \rightarrow\hspace{-4pt}+\hspace{3pt} b$~~and	& For $b$ to occur, $a$ must occur exactly once prior \\
									& $b \rightarrow\hspace{-4pt}\%\hspace{3pt} b$ 		& \\
\chainprecedence$(a, b)$	& \footnotemark										& For $b$ to occur, $a$ must occur immediately prior \\
\notchainsuccession$(a, b)$	& $a\rightarrow\hspace{-4pt}\%\hspace{3pt} b$		& Activity $b$ may not occur immediately after $a$ \\
\notcoexistence$(a, b)$		& $a \rightarrow \hspace{-4pt} \% \hspace{3pt} b \land b \rightarrow \hspace{-4pt} \% \hspace{3pt} a$ & Activities $a$ and $b$ may not co-occur in the same trace \\
\bottomrule
\end{tabu}
\label{tab:parnek_core}
\caption{Relevant constraint templates from Declare.}
\end{table*}
\footnotetext{The \chainprecedence~ relation is not straightforward to encode in DCR Graphs relations and in fact, ParNek looks for evidence of \chainprecedence relations, but encodes them as $a \rightarrow\hspace{-3pt}+\hspace{3pt} b, b\rightarrow\hspace{-3pt}\%\hspace{3pt} b$}

\paragraph{\textbf{Initialization} (lines: \ref{alg:initstart}-\ref{alg:initfinish})}
We begin by defining a set of events 
$$E \equiv \{0,\ldots, |\Sigma_L|\}$$
containing the same number of events as distinct activities present in the log, the latter defining our set of activities
$$A \equiv \Sigma_L.$$
The labeling function 
$$l : E \rightarrow \Sigma_L;~i \mapsto s_i$$
is a bijective mapping between events and activities. So for all intents, events and activities are equivalent. 
Finally we assign an initial marking 
$$m \equiv (E, \emptyset, \emptyset)$$ in which all events are included, none are pending, and none are executed.

\paragraph{\textbf{Self-Exclusions} - \atmostone~(line: \ref{alg:selfex}):} We begin with activities for which the log satisfies the \atmostone~ relation. Any activity $s$ satisfying this unary relation are mapped onto the binary self-exclusion relation $s \rightarrow \hspace{-4pt} \% \hspace{3pt} s$.

\paragraph{\textbf{Responses} - \responsedeclare~(line: \ref{alg:response}):} All pairs of \emph{distinct} activities $s$ and $t$ for which the log satisfies the \responsedeclare~ relation, are mapped directly onto the response relation $s \hspace{2pt} \bullet \hspace{-4pt} \rightarrow t$.

\paragraph{\textbf{Conditions} - \precedence~(line: \ref{alg:precedence}):} All pairs of \emph{distinct} activities $s$ and $t$ for which the log satisfies the \precedence~ relation, are mapped directly onto the condition relation $s \hspace{2pt} \bullet \hspace{-4pt} \rightarrow t$. While this forms the basis of the condition relation, more will be added in lines \ref{alg:morecondstart}-\ref{alg:morecondfinish}. 

\paragraph{\textbf{Includes/Excludes} - \chainprecedence~(line: \ref{alg:altprec1}-\ref{alg:altprec2}):} The first step in populating $+$ and adding further self-exclusions to $\%$, is based on identifying \chainprecedence~ relations. However, encoding \chainprecedence~ in DCR Graphs is less straightforward than \alternateprecedence, which is (nearly\footnote{In order to completely capture \alternateprecedence, the target activity needs to be excluded in the initial marking. This can lead to complications w.r.t. other relations in which the target is source, and is therefore omitted.}) captured by an include and self-excludes. Since \alternateprecedence~ \emph{subsumes} \chainprecedence, it is safe to check for evidence of the more restricted \chainprecedence, yet add \alternateprecedence~ to the model.

\paragraph{\textbf{Excludes - Predecessor/Successor} (lines: \ref{alg:notcoexistence1}-\ref{alg:notsuccession3}):}
Further excludes relations are found by defining two relations:
$$
Predecessor(L) \text{ and } Successor(L)
$$
which return the sets of \emph{all possible} predecessors and successors of an activity, respectively.

Based on the observation that a log in which activities $s$ and $t$ never co-occur in the same trace satisfies the \notcoexistence$(s,t)$~ relation, we add $s \rightarrow \hspace{-4pt} \% \hspace{3pt} t$ and $t \rightarrow \hspace{-4pt} \% \hspace{3pt} s$ (lines: \ref{alg:notcoexistence1}-\ref{alg:notcoexistence2}). However, due to the subsequent removal of redundant exclusions (lines: \ref{alg:removeex1}-\ref{alg:removeex2}), the \notcoexistence~ relation cannot be guaranteed to hold since one or both of the exclusions may be removed. 

Furthermore, if $s$ is observed to precede, but never succeed $t$, and if no self-exclusion $s \rightarrow \hspace{-4pt} \% \hspace{3pt} s$ has been found, we add $t \rightarrow \hspace{-4pt} \% \hspace{3pt} s$ (lines: \ref{alg:notsuccession1}-\ref{alg:notsuccession3}). 

In order to restrain model complexity, only one exclusion relation is included for each target activity by means of the $ChooseOneRelation$ function. At present, this function is implemented in a first-come manner with a more sophisticated approach being left for future work.

\paragraph{\textbf{Includes \& Excludes} - \notchainsuccession~(lines: \ref{alg:notchainsuccession1}-\ref{alg:notchainsuccession2}):} To identify further includes and excludes relations, we rely on $NotChainSuccession(L)$ as well as $Between(L)$, which simply identifies activities occurring between two other activities in a log.

Put simply, if we never observe $s$ followed immediately by $t$, we add an exclusion $s \rightarrow \hspace{-4pt} \% \hspace{3pt} t$ (\notchainsuccession). If however, $t$ occurs after $s$, with some sequence of intermediate activities s.t. we have $\langle \ldots, s, u_1, \ldots, u_n, t, \ldots \rangle$, then we allow all intermediate events to re-include $t$. That is, for all $1 \leq i \leq n$, we add $u_i \rightarrow \hspace{-4pt} + \hspace{3pt} t$.

\paragraph{\textbf{Remove Redundant Excludes} (lines: \ref{alg:removeex1}-\ref{alg:removeex2}):} Based on the observation that if activity $r$ always precedes $s$, and if\\ $r\rightarrow\hspace{-4pt}\%\hspace{2pt} t$, then adding $s \rightarrow\hspace{-4pt}\%\hspace{2pt} t$ is redundant. It should be noted that this redundancy does not hold if some $u$ occurs between $r$ and $s$ and $u \rightarrow\hspace{-3pt}+\hspace{3pt} t$. Presently, this caveat is ignored, potentially leading to a decrease in model precision, but allowing for an enormous reduction in model complexity.

\paragraph{\textbf{Transitive Reduction} (lines: \ref{alg:transred1}-\ref{alg:transred2} and \ref{alg:transred3}):} The condition and response relations satisfy the transitive property when seen in isolation. That is, if we have $s \condition t$ and $t \condition u$, then $s \condition u$. In this case, $s \condition u$ is superfluous. The caveat, \emph{seen in isolation}, is crucial however, since if the same model has $v \rightarrow \hspace{-4pt} \% \hspace{3pt} t$ for some $v$, then $t$ may become excluded, annulling the implicit $s \condition u$. Formally,

$$
s \condition t \land t \condition u \land \nexists v.~v \rightarrow \hspace{-4pt} \% \hspace{3pt} t \models s \condition u
$$

In fact, we can safely remove redundant $s \condition u\hspace{3pt}$ despite the presence of an interfering excludes relation (that is, we ignore $\nexists v.~v \rightarrow \hspace{-4pt} \% \hspace{3pt} t$). The removal is safe in the sense that this can only result in a more permissive model, i.e.~we do not risk arriving at a model on which the log cannot be replayed. The downside is a less precise model, which may permit behavior which ought to be forbidden. 

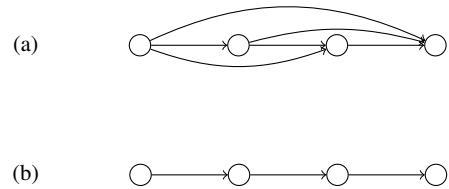
\begin{figure}[!h]
	\begin{center}
		\begin{tikzpicture}
			\begin{scope}
			    \node [circle] (l1) {(a)};
    			\node [draw, circle, right=of l1] (n1) {};
    			\node [draw, circle, right=of n1] (n2) {};
    			\node [draw, circle, right=of n2] (n3) {};
    			\node [draw, circle, right=of n3] (n4) {};
    			
    			\path   (n1) edge[->] (n2)
    			        (n2) edge[->] (n3)
    			        (n3) edge[->] (n4)
    			        (n1) edge[->, bend right=20pt] (n3)
    			        (n1) edge[->, bend left=25pt] (n4)
    			        (n2) edge[->, bend left=15pt] (n4);
    			        
    			\node [circle, below=of l1] (l2) {(b)};
    			\node [draw, circle, right=of l2] (n5) {};
    			\node [draw, circle, right=of n5] (n6) {};
    			\node [draw, circle, right=of n6] (n7) {};
    			\node [draw, circle, right=of n7] (n8) {};
    			
    			\path   (n5) edge[->] (n6)
    			        (n6) edge[->] (n7)
    			        (n7) edge[->] (n8);
			\end{scope}	
		\end{tikzpicture}
	\end{center}
	\caption{Transitive reduction: graph (a) has the same reachability/transitive closure as the reduced graph (b).}
	\label{fig:transred}
\end{figure}

Transitive reduction is performed on all condition and response relations prior to the final step of discovering additional condition relations, and once again on condition relations afterwards. In many models the reduction in relations is very substantial. See Figure \ref{fig:transred} for a graphical illustration.

\paragraph{\textbf{Additional Conditions} (lines: \ref{alg:morecondstart}-\ref{alg:morecondfinish}):} The first set of conditions we added based on the $\textsc{Precedence}$ relations were conservative in that this relation was observed to hold unconditionally across traces. We can now add less obvious conditon relations, taking advantage of semantics added to our model by inclusion and exclusion relations.

We start by adding $s \hspace{-2pt}\rightarrow\hspace{-5pt}\bullet\hspace{3pt} t$ if $s$ occurs before the \emph{first} occurrence of $t$ in \emph{some} trace. For those traces in which $s$ does not precede the first $t$, it may be the case that $s$ is excluded by some other activity $u$ if the relation $u \rightarrow \hspace{-4pt} \% \hspace{3pt} s$ is present and $u$ is observed prior to $t$. Recall that DCR Graph semantics dictate that a relation does not apply when the source activity is excluded.

Since only includes and excludes relations are determinative for the validity of these candidate relations, we can utilize a limited replay strategy based on these relations alone. This approach is less computationally demanding than using the full model.

\begin{table*}[h!]
\begin{tabularx}{\textwidth}{X r | X r}
$\boldsymbol{AtMostOne}: \mathcal{L} \rightarrow 2^\Sigma;$ 					& 														& \multicolumn{2}{l}{$\boldsymbol{ChooseOneRelation}: 2^{S \times S} \rightarrow S \times S;$} \\
$L \mapsto \{~\sigma(i)~|$ 														& $\forall \sigma \in L.$								& \multicolumn{2}{l}{$\{(s_1,t_1),(s_2,t_2),\ldots,(s_n,t_n)\} \mapsto (s_i,t_i) \text{~where~} 1 \leq i \leq n$}\\
																				& $\sigma(i) = \sigma(j)$								& & \\
																				& $\implies \exists j.~t = \sigma(j) \land i < j~\}$ 	& & \\
$\boldsymbol{Response} : \mathcal{L} \rightarrow 2^{\Sigma \times \Sigma};$ 	& 														& $\boldsymbol{Inclusions}: \mathcal{L} \rightarrow 2^{\Sigma \times \Sigma};$ 	& \\
$L \mapsto \{~(s, t) \in \Sigma_L \times \Sigma_L~|$							& $\forall \sigma \in L.$								& $L \mapsto \{~(s, t) \in \Sigma_L \times \Sigma_L~|$ 							& $\forall \sigma \in L.~s = \sigma(i)$ \\
																				& $s = \sigma(i)$										&																				& $\implies~\exists j.~t = \sigma(j)$ \\ 
																				& $\implies \exists j.~t = \sigma(j) \land i < j~\}$ 	&																				& $\land~i = j-1~\}$ \\
$\boldsymbol{Precedence} : \mathcal{L} \rightarrow 2^{\Sigma \times \Sigma};$ 	&														& $\boldsymbol{Predecessors} : \mathcal{L} \rightarrow 2^{\Sigma \times \Sigma};$ & \\ 
$L \mapsto \{~(s, t) \in \Sigma_L \times \Sigma_L~|$							& $\forall \sigma \in L.$								& $L \mapsto \{~(s, t) \in \Sigma_L \times \Sigma_L~|$							& $\exists \sigma \in L.~t = \sigma(j)$ \\ 
																				& $t = \sigma(j)$										&																				& $\implies~\exists i.~s = \sigma(i)$\\
																				& $\implies \exists i.~s = \sigma(i) \land i < j~\}$	& 																				& $\land~i < j~\}$ \\
\multicolumn{2}{l|}{$\boldsymbol{AlternatePrecedence} : \mathcal{L} \rightarrow 2^{\Sigma \times \Sigma};$}								& $\boldsymbol{Successors} : \mathcal{L} \rightarrow 2^{\Sigma \times \Sigma};$	& \\
$L \mapsto \{~(s, t) \in \Sigma_L \times \Sigma_L~|$							& $\forall \sigma \in L.$								& $L \mapsto \{~(s, t) \in \Sigma_L \times \Sigma_L~|$							& $\exists \sigma \in L.~s = \sigma(i)$ \\
																				& $t = \sigma(j)$										& 																				& $\implies~\exists j.~t = \sigma(j)$ \\
																				& $\implies \exists i.~s = \sigma(i) \land i < j~\land$	& 																				& $\land~i < j~\}$\\					
																				& $\nexists k.~t = \sigma(k) \land i < k < j~\}$		& 																				& \\													
\multicolumn{2}{l|}{$\boldsymbol{ChainPrecedence} : \mathcal{L} \rightarrow 2^{\Sigma \times \Sigma};$}									& $\boldsymbol{Between} : \mathcal{L} \rightarrow 2^{\Sigma \times \Sigma};$	& \\
$L \mapsto \{~(s, t) \in \Sigma_L \times \Sigma_L~|$							& $\forall \sigma \in L.$								& $L \mapsto \{~u  \in \Sigma_L~|$												& $\nexists \sigma \in L.$ \\
																				& $t = \sigma(j)$										& 																				& $s = \sigma(i)~\land$ \\
																				& $\implies \exists i.~s = \sigma(i)$					& 																				& $t = \sigma(j)~\land$ \\
																				& $\land~i = j-1~\}$									& 																				& $u = \sigma(k)~\land$ \\
																				&														&																				& $s \neq u \land s \neq u$\\
																				&														&																				& $\implies~i < k < j ~\}$ \\
\multicolumn{2}{l|}{$\boldsymbol{NotChainSuccession} : \mathcal{L} \rightarrow 2^{\Sigma \times \Sigma};$}								& \multicolumn{2}{l}{$\boldsymbol{TransitiveReduction} : 2^{\Sigma \times \Sigma} \rightarrow 2^{\Sigma \times \Sigma};$} \\ 
$L \mapsto \{~(s, t) \in \Sigma_L \times \Sigma_L~|$							& $\nexists \sigma \in L.$								& $R \mapsto \{R' \subseteq ~R~|$												& $\forall (a,c) \in R',$ \\													 																				& $s = \sigma(i)~\land$									&																				& $(a,c) \in R ~\land\nexists b.$\\
																				& $t = \sigma(j)~\land$									&																				& $(a,b) \in R'$ \\
																				& $\implies~i \neq j-1~\}$									&																				& $\land~(b,c) \in R'~\}$ \\																																								 																																
\end{tabularx}
\label{tab:functions}
\caption{Formal definitions of helper functions which return sets of relevant relations. All functions have event logs as their domain ($\mathcal{L}$), except $ChooseOneRelation$ and $TransitiveReduction$.}
\end{table*}

\begin{algorithm*}
\SetArgSty{}
\normalsize
\SetKwInOut{Input}{input}\SetKwInOut{Output}{output\hspace{0.25pt}}
\LinesNumbered
\SetAlgoLined
\DontPrintSemicolon
\setlength{\algomargin}{2em}
\Input{A log $L$}
\Output{A DCR Graph $G$}

\BlankLine
\BlankLine
\tcp*{INITIALIZATION}
$E \equiv \{0,\ldots, |\Sigma_L|\}$ \tcp*{ set of events} \label{alg:initstart}
$A \equiv \Sigma_L$ \tcp*{ activities}
$l~~\equiv i \in E \mapsto s_i \in \Sigma_L$ \tcp*{ bijective labeling}
$m \equiv (E, \{\}, \{\})$ \tcp*{ initial marking} \label{alg:initfinish}
\BlankLine
\BlankLine
\BlankLine
\tcp*{DECLARE TEMPLATES}
$\% \hspace{10pt} \equiv AtMostOne(L) \times AtMostOne(L)$ \tcp*{self exclusions} \label{alg:selfex}
$\response\hspace{6pt} \equiv Response(L)$ \tcp*{response relations} \label{alg:response}
$\condition\hspace{3pt} \equiv Precedence(L)$ \tcp*{condition relations} \label{alg:precedence}
$+ \hspace{10pt} \equiv + \cup \{~(s,t)~|~\forall s, s\neq t.~(s,t) \in ChainPrecedence(L)~\}$ \tcp*{alternate precedence} \label{alg:altprec1}
$\% \hspace{7pt} := \% \cup \{~(t,t)~|~\exists s, s\neq t.~(s,t) \in ChainPrecedence(L)~\}$ \tcp*{alternate precedence} \label{alg:altprec2}
\BlankLine
\BlankLine
\BlankLine
%\tcp*{ choose method for defining in/exclusions}
%\eIf{naive\_inex}{
%$+ \leftarrow Inclusions(L)$\;
%$\% \leftarrow \% \cup +^C$  
%}{
%\begin{align*}
\BlankLine
\BlankLine
\BlankLine
\tcp*{ADDITIONAL EXCLUDES}
$\% := \% \cup ChooseOneRelation\big(~\{~(s,t)~|~\forall s, s\neq t.~(s,t) \notin Predecessors(L)~\land$ \tcp*{not coexistence} \label{alg:notcoexistence1}
$\hspace{218pt}(t,s) \notin Successors(L)~\}~\big)$\; \label{alg:notcoexistence2}
%\end{align*}
%\begin{align*}
$\% := \% \cup ChooseOneRelation\big(~\{~(t,s)~|~\forall s, s\neq t.~(s,t) \in Predecessors(L)~\land$ \tcp*{not succession} \label{alg:notsuccession1}
$\hspace{218pt}(t,s) \notin Successors(L)~\land$\; \label{alg:notsuccession2}
$\hspace{218pt}(s,s) \notin \%\}~\big)$\; \label{alg:notsuccession3}
%\end{align*}
%}
\BlankLine
\BlankLine
\BlankLine
\tcp*{ADDITIONAL INCLUDES/EXCLUDES}
%\If{improve}{
$\% := \% \cup \{~(s,t)~|~(s,t) \in NotChainSuccession(L)~\}$ \tcp*{not chain succession} \label{alg:notchainsuccession1}
$+ := + \cup \{~(u,t)~|~\exists s.~(s,t) \in NotChainSuccession(L) \land (s,u,t) \in Between(L)~\}$\;\label{alg:notchainsuccession2}
%\If{optimize}{ 
\BlankLine
\BlankLine
\BlankLine
\tcp*{REMOVE 'REDUNDANT' EXCLUSIONS}
%	\begin{align*}
$\% := \% \setminus \{~(s,t)~|~\exists u.~(u,t) \in \%~\land$\; \label{alg:removeex1}
$\hspace{94pt}(u,s) \in AlternatePrecedence(L)$\; \label{alg:removeex2}
%	\end{align*}
%	}
%}
\BlankLine
\BlankLine
\BlankLine
%\If{reduce}{
\tcp*{REMOVE 'REDUNDANT' CONDITIONS/RESPONSES}
$\response\hspace{6pt} := TransitiveReduction(\response\hspace{3pt})$\; \label{alg:transred1}
$\condition\hspace{3pt} := TransitiveReduction(\condition\hspace{3pt})$\; \label{alg:transred2}
%}
\BlankLine
\BlankLine
\BlankLine
\tcp*{ADDITIONAL CONDITIONS}
%\If{more\_conds}{
%\begin{align*}
$\condition\hspace{3pt} := \hspace{5pt} \condition \cup \{~(s,t)~|~(\exists \sigma \in L.\forall k.~s=\sigma(i) \land t=\sigma(j)=\sigma(k)~\land~i < j \leq k~)~\land$\; \label{alg:morecondstart}
$\hspace{98pt}(\forall \sigma \in L.\forall i>j.~s=\sigma(i) \land t=\sigma(j)~\land$\;
$\hspace{165pt}\exists h < j.~r=\sigma(h) \land r \rightarrow\hspace{-3pt}\%\hspace{3pt} s) \}$\;\label{alg:morecondfinish}
%\end{align*}
%}
\BlankLine
\BlankLine
\BlankLine
$\condition\hspace{3pt} := TransitiveReduction(\condition\hspace{3pt})$\; \label{alg:transred3}
\BlankLine
\BlankLine
\BlankLine
$\Return~(E, M, A, \response, \condition, +, \%, l)$ \tcp*{RETURN DCR GRAPH}
\caption{High-level control flow of the mining algorithm.\label{ParNek}}
\end{algorithm*}

\section{The DisCoveR Miner}
\label{sec:implementation}

\lstset{language=Java,
   showstringspaces=false,
    basicstyle=\small\ttfamily,
    breaklines=true,
    keywordstyle=\color{blue},
    commentstyle=\color[gray]{0.6},
    stringstyle=\color[RGB]{255,150,75},
    tabsize=3
}

%basicstyle=\small\ttfamily,columns=fullflexible

%Two implementations of ParNek exist. The first was developed as part of the M
%ParNek was implemented …. 
DisCoveR is a fully open source (licensed under LGPL-3.0) JAVA implementation of the ParNek algorithm. It was developed as an alternative to the original implementation which had stricter licensing terms. 
The relaxed licensing has enabled the straightforward integration of the algorithm in industrial solutions and allows for the development of extensions to the algorithm without these falling under an overly restrictive license.

In addition to these licensing advantages, the DisCoveR algorithm also offers improved performance by using a highly efficient implementation of DCR Graphs inspired by earlier work by Debois et al.~\cite{debois_dcr_2017,madsen2018collaboration}. In this implementation the relations and markings of DCR graphs are represented as bit vectors, each activity corresponding to a particular index of the vectors. For example, the marking can be represented as such:
\begin{lstlisting}
public BitSet executed = new BitSet();
public BitSet included = new BitSet();
public BitSet pending = new BitSet();
\end{lstlisting}

And relations as such:
\begin{lstlisting}
public HashMap<Integer, BitSet> 
    conditionsFor = new HashMap<>();
public HashMap<Integer, BitSet> 
    responsesTo = new HashMap<>();
public HashMap<Integer, BitSet> 
    excludesTo = new HashMap<>();
public HashMap<Integer, BitSet> 
    includesTo = new HashMap<>();
\end{lstlisting}

The semantics can then be expressed as a short list of bitvector operations, in particular, enabledness of events can be computed as follows:

\begin{lstlisting}
public Boolean enabled(final BitDCRMarking marking, final int event) {
    // The event is not included.
    if (!marking.included.get(event))
    	return false;
    // Any of the conditions for the event are included and have not been executed.
    if (conditionsFor.get(event).intersects(marking.blockCond()))
    	return false;
    return true;
}

// Method on the class BitDCRMarking 
public BitSet blockCond() {
	return included.clone().andNot(executed);
}
\end{lstlisting}

Note that BitSets are JAVA's version of bit vectors, the get() method retrieves the bit at a given index, the intersects method is essentially applies an AND operation on two vectors and checks if the result is 0.
The execution of an event can be computed as follows:
\begin{lstlisting}[escapeinside={(*}{*)}]
public BitDCRMarking execute(final BitDCRMarking marking, final int event) {
    // Copy the previous marking
    BitDCRMarking result = marking.clone();
    // Set the event as executed
    result.executed.set(event);
    // Clear the event as no longer pending
    result.pending.clear(event);
    // Add all new pending responses
    result.pending.or(responsesTo.get(event));
    // Exclude excluded events
    result.included.andNot(excludesTo.get(event));
    // Include included events
    result.included.or(includesTo.get(event));
    return result;
}
\end{lstlisting}
%(*\small\ttfamily\textcolor{gray} {\ref{def:dcr}}*)
 % mathescape=true

This implementation of DCR Graphs allows for extremely fast replay of logs, which significantly reduces the duration of the \emph{Additional Conditions} part of the algorithm, which requires a replay of the log on the graph that has been found up-to that point.

Furthermore, to avoid repeating computations, we separate the mining process into two steps: first we build a number of relevant abstractions of the log, which we then use afterwards during the actual model building steps as described in Section~\ref{sec:algorithm}. 
This separation of concerns ensures that there is a central part of the code where we parse the log, with all other parts of the algorithm working only on these abstractions, which are bounded by the number of activities and not the log size.
Inspired by the efficient implementation of DCR Graphs, we also store and compute these abstractions through bit vector operations. The listing below shows their definition:
\begin{lstlisting}
public HashMap<Integer, BitSet> chainPrecedenceFor = new HashMap<>();
public HashMap<Integer, BitSet> precedenceFor = new HashMap<>();
public HashMap<Integer, BitSet> responseTo = new HashMap<>();
public HashMap<Integer, BitSet> predecessor = new HashMap<>();
public HashMap<Integer, BitSet> successor = new HashMap<>();
public BitSet atMostOnce = new BitSet();
\end{lstlisting}

The listing below shows how some of these abstractions are computed. For brevity's sake we show only some of the simpler abstractions to compute, we note however that all abstractions can be computed in linear time, i.e. none of them requires a nested iteration over the log or current trace. For convenience, logs are transformed into lists of integers, this allows for straightforward mapping of activities to the indices of the bit vectors and efficient storage of the log for later reuse.
%\begin{minipage}{\linewidth}
\begin{lstlisting}
public void parseTrace(List<Integer> t) {
    // A helper set to keep track of which activities were seen at least once before in this trace.
	BitSet localAtLeastOnce = new BitSet();
	for (int i : t) {
	    // Any activities that were seen at least once before i ware predecessors for i
		predecessor.get(i).or(localAtLeastOnce);
		// If i was seen before in the trace, then it occurs more than once.
		if (localAtLeastOnce.get(i))
			atMostOnce.clear(i);
		// Add the current activity to those seen at least once
		localAtLeastOnce.set(i);
		// for there to be a precedence relation between i and an activity, it needs to have happened before i in all traces.
		precedenceFor.get(i).and(localAtLeastOnce);
	}
}
\end{lstlisting}
%\end{minipage}

Altogether, these optimizations provide us with an extremely efficient implementation of the ParNek algorithm. In the following section we will show through experimentation that it is in fact one order of magnitude faster than any other DCR Graphs miner and two orders of magnitude faster than the state-of-the-art in Declare mining.

\section{Evaluation}
\label{sec:evaluation}

To evaluate the performance of our algorithm, we frame the process discovery task as a binary classification task of identifying legal/illegal traces. For this, we take advantage of a labeled dataset from the Process Discovery Contest 2019 \footnote{\url{https://icpmconference.org/2019/process-discovery-contest}}, in which \discover~was the second-best performing algorithm, classifying $96.1\%$ of traces correctly. This result was achieved despite that \discover~considers only control-flow, ignoring auxillary data associated with events.

For comparison, we also report results for an existing DCR Graph miner \cite{debois2017declarative}, which uses a greedy strategy for discovering relations, as well as the MINERful miner, the state-of-the-art miner for Declare \cite{ciccio2015discovery}.

Framing process discovery as a binary classification task is arguably an oversimplification of the aim of process discovery, since it does not capture the \emph{degree} to which a model fails to capture an event log. Error measures that aim to capture this are usually based on model-log \emph{alignment} techniques \cite{adriansyah2012alignment}, or model specific measures such as \emph{token replay} metrics for Petri nets \cite{rozinat2008conformance}. The advantage of the classification formulation lies in the ease of interpretability and comparability. In a model-agnostic manner, we gain a view of the algorithm's bias towards committing different classes of statistical errors (e.g. Type I/II) by analyzing \emph{true/false positives/negatives}, and the corresponding \emph{precision},  \emph{recall}, \emph{$F_1$}-score and MCC measures.

Before presenting the results, we briefly formalize the task of learning and evaluation of a classifier in the context of process discovery.

\subsection{The Learning Task}

The goal of a classification task is to learn an approximation $h$ of a \emph{target function} $f$ which is assumed to generate the observed data \cite{yaser2012learning}. The training data $L$ is an i.i.d. sample from the true probability distribution ($\mathbb{P}_P$) associated with $f$. The aim is to maximize performance (e.g. minimize an error function) on \emph{out-of-sample} data by means of optimizing performance on \emph{in-sample} training data in such a way that the learned model avoids overfitting. 

Formally, a learning algorithm $\gamma$ is a mapping from a sampling $L$ from the process $(P, \mathbb{P}_P)$ to a hypothesis space $\mathcal{H}$ s.t. the \emph{out-of-sample} error $E_{out}$ is minimized:

$$
\gamma : \mathcal{L} \rightarrow \mathcal{H}; ~L \mapsto \argmin_{h\in \mathcal{H}}E_{out}(h)
$$

To define our error function $E$, we can frame process discovery-based binary classification as the task of predicting the outcome of a random Bernoulli variable defined by

$$\mathbbm{1}(\sigma \in \ell(q))$$ 
which returns 1 when a trace $\sigma$ is a member of the language of model $q$, and 0 otherwise.

The most straightforward way of defining the \emph{in-sample} error measure, is simply the proportion of ``successes'' in this Bernoulli trial:

$$
E_{in}(h) = \sum_{\sigma \in L} \frac{\mathbbm{1}(\sigma \in \ell(h))}{|L|}  
$$ 

In this formulation, the \emph{out-of-sample} error follows directly from our definition of the underlying distribution and simply represents the probability of sampling a trace from the target function (true process) which is rejected by $h$:

$$
E_{out}(h) = \mathbb{P}_P(\sigma \notin \ell(h))
$$

However, typically the application domain will call for a more nuanced error measure which accounts more precisely for the \emph{type} of error a classifier makes.

This can be quantified by distinguishing between type I and type II errors with user specified penalties:

$$
E_\sigma(h,f) = \begin{cases} 
0 & \text{ if } \sigma \in \ell(h) \land \sigma \in P \\
\alpha & \text{ if } \sigma \in \ell(h) \land \sigma \notin P \qquad \text{type I}\\
\beta & \text{ if } \sigma \notin \ell(h) \land \sigma \in P \qquad \text{type II}
\end{cases}
$$

For example, in a high-security setting, a false positive could mean allowing an intruder entry or failing to identify fraudulent behavior. In such a scenario, the penalty $\alpha$ for a false positive should greatly outweigh the penalty $\beta$ for the inconvenience of incorrectly denying entry or auditing a compliant case.

\paragraph{Regularization} Minimizing $E_{in}$ is an almost trivial task given a large enough hypothesis space $\mathcal{H}$, since a model can be found which fits the in-sample data nearly exactly. However, such a model will almost certainly fail to generalize to out-of-sample data. This is because, while a large enough $\mathcal{H}$ may indeed contain the target function $f$, the likelihood of our learning algorithm choosing $f$ in such a large hypothesis space is vanishingly small. It is much more likely to settle on some other, very complex, function $g \in \mathcal{H}$, leading to a high $E_{out}$. While counter-intuitive, restricting $\mathcal{H}$ to a smaller set which does not include $f$ will often lead to a lower $E_{out}$.  

Thus, a key component in the learning process is that of \emph{regularization}: a process for controlling the complexity of a learned model, i.e.~ restricting the size of the hypothesis space, to improve generalization. This gives rise to the formulation of the learning process as a trade-off between inductive \emph{bias}\footnote{The minimal in-sample error achievable for hypothesis $h \in \mathcal{H}$.} of a hypothesis set and a penalty for the \emph{complexity} of a hypothesis \cite{shalev2015understanding}. The sum of these terms gives an estimate of the out-of-sample error:

$$
\hat{E}_{out} = E_{in} + \Omega(N, \mathcal{H}, \delta).
$$
Where $N$ denotes sample size, $\mathcal{H}$ the hypothesis space and $\delta$ the desired confidence that $\hat{E}_{out}\leq E_{out}$.

So although we can achieve a very low in-sample error using a rich hypothesis set, we penalize complex models using a \emph{regularization} function $\Omega$. Explicitly incorporating this function into learning algorithms s.t.~it minimizes $\hat{E}_{out}$ rather than $E_{in}$, can greatly improve results.

ParNek does not currently attempt to explicitly minimize $\hat{E}_{out}$, and $\Omega$ is likewise not explicitly formulated. However, some form of regularization is achieved by effectively restricting the size of $\mathcal{H}$. This is done via a set of heuristics attempting to control model complexity, removing those which are redundant w.r.t. training data or add little to the precision of its semantics. Indeed, ParNek cannot discover the entire set of DCR Graphs, thus

$$
\mathcal{H}_{ParNek} \subset \mathcal{H}_{DCR} = \omega\text{-regular languages}
$$
Restricting the available hypothesis set is analogous to limiting a linear regression algorithm to third-order polynomials, for example, which corresponds to an $\Omega$ which assigns a zero weight to all higher-order coefficients. 

While heuristic in nature, the approach is effective, as is clearly seen in comparison to other miners which do little to control model complexity, such as \baselineminer. We intend to pursue more well-defined regularization procedures for DCR Graph mining algorithms in future work.

\paragraph{Other Metrics} Aggregate evaluation metrics, such as \emph{precision}, \emph{recall} and \emph{$F_1$-score} are commonly reported for classification tasks. Given a confusion matrix, we define precision(prec.) and recall as follows: 

\begin{table}[!h]
\begin{tabularx}{0.5\textwidth}{c | c c | c }
\textsc{Pred-}	& \multicolumn{2}{c|}{\textsc{Data}} & \\
\textsc{iction}	& $+$ 				& $-$			& \\
\midrule
$+$				& True Pos.($TP$)	& False Pos.($FP$)& prec. $\equiv \frac{TP}{TP + FP}$\\
$-$				& False Neg.($FN$)	& True Neg.($TN$) & \\
\midrule
				& recall $\equiv \frac{TP}{TP + FN}$ & & \\
\end{tabularx}
\end{table}

The $F_\beta$-score is then the harmonic mean of precision and recall, where $\beta$ determines a weighting of precision relative to recall:

$$
F_{\beta} = \frac{(1+\beta^2) \cdot \text{precision} \cdot \text{recall}}{\beta \cdot \text{precision} + \text{recall}}
$$

Originally stemming from information retrieval, these metrics have been criticized for giving weight to true positives and ignoring true negatives \cite{chicco2020advantages}, and other metrics such as Matthews Correlation Coefficient (MCC) avoid assumptions regarding the target class. 

Arguably, process mining \emph{can} be seen as an information retrieval task, if the tool is used to ``query'' an event log for compliant/noncompliant traces. For completeness, we report precision, recall and $F_1$-score for both the situation in which the target class is compliant behavior (true positive) and noncompliance (true negative), as well as Matthews Correlation Coefficient (MCC).

\subsection{Results}
\label{sec:results}

In addition to case studies, we present a controlled evaluation of the algorithm based on a labeled data set from the Process Discovery Contest 2019 \footnote{\url{https://icpmconference.org/2019/process-discovery-contest}}. The evaluation is bolstered by the truly blind nature of the process. After being presented with an \emph{unlabeled} training set and submitting results for a partially blind validation round, the predictions on a separate test set were sent in to the contest administrators who independently evaluated their accuracy. This removes any potential for accidental data snooping.

See Table \ref{tab:confusion_matrices} for the complete results.

\paragraph{Dataset} The data set essentially consists of 10 independent data sets stemming from 10 different processes. Participants were presented with an unlabeled training set from each process. Then, two validation sets were provided for which participants could submit their algorithm's classification results. The organizers then returned a confusion matrix - but no details regarding which traces specifically were misclassified and how. Two rounds of submission for validation were permitted, though we only took advantage of the first.

Event logs for processes 1, 5, 7, 8, 9, and 10 contained auxiliary data associated with each event, sometimes more than one attribute. The version of our algorithm presented here considers only control-flow and is unable to take advantage of additional attributes, and neither do the miners we present in the following comparison.

\paragraph{Comparison} For comparison, we present the performance of two similar mining algorithms, the first is another DCR Graph mining algorithm designed by Debois, et al~\cite{debois2017declarative}. The second, is the state-of-the-art among miners based on Declare constraints, \minerful~\cite{ciccio2015discovery}; 

Debois, et al's DCR Graph miner takes a very greedy approach to identifying DCR relations which hold for an event log. Essentially, the algorithm begins with a fully constrained model over the set of activities in the log (mapped one-to-one to DCR events), then goes through the log and removes any constraints which are violated by observed behavior. 

Due to the greedy strategy, the algorithm often finds thousands of constraints and clearly overfits the training data, leading to poor performance on test data.

\minerful~is a sophisticated miner for the Declare language which uses a number of user-defined parameters to determine which constraints to include in a model after mining the event log. The three core parameters are:

\begin{labeling}{\textsc{Interest Factor}}
  \item [Support] The fraction of traces in which the constraints must hold.
  \item [Confidence] Support scaled by the fraction of traces in which a constraint is activated.
  \item [Interest Factor] Confidence scaled by the fraction of traces in which target of a constraint is also present.
\end{labeling}

A constraint is considered to be \emph{activated} when it becomes relevant in a trace. So, a succession constraint between $s$ and $t$ will only become activated in traces in which $s$ is present. In additional, to count towards interest factor, the target $t$ must also be present. Defined as scalings, these parameters are dependent on one another and result in the bounds: support > confidence > interest factor.

\minerful~also performs subsumption checks to eliminate redundant or meaningless constraints. For example, wherever a \chainsuccession~constraint is found to hold, \succession~will necessarily hold and adds no information. This procedure is akin to \discover's strategy of removing transitively redundant constraints in order to avoid unnecessarily complex models.

In our comparison, we held the support threshold fixed at 1.0 for comparability with \discover~and \baselineminer~which both guarantee perfectly fitting models. For confidence and interest factor, we employed an automated parametrization procedure originally developed for the evaluation in \cite{back2018towards}. The procedure employs a binary search strategy to find values for confidence and threshold which result in a model with a number of constraints as close to, but not exceeding, some limit. We present results for models with between 89 and 500 constraints, which encompasses the range of model sizes generated by \discover.

\paragraph{Results} We report results for the classification task in a confusion matrix for each of the 10 processes, as well as aggregate across processes in Table \ref{tab:confusion_matrices}. Keep in mind, that a user-defined error measure may choose to weigh false positives and false negatives differently ($\alpha$ and $\beta$ in our formalization).

Additionally, we report Matthews Correlation Coefficient (MCC) in addition to precision, recall, and $F_1$-score, both in the case of the target class being permissible traces, as well as forbidden traces. The appropriate framing would depend on the application.    

\subsubsection{Run-time}

\discover~ outperforms quite markedly in terms of run-time. We compare performance to the same two miners in our classification evaluation: \baselineminer~and \minerful. We find that \discover~performs an \emph{order of magnitude} better than \baselineminer~and nearly two orders of magnitude better than \minerful.

\begin{figure}
\includegraphics[width=0.5\textwidth]{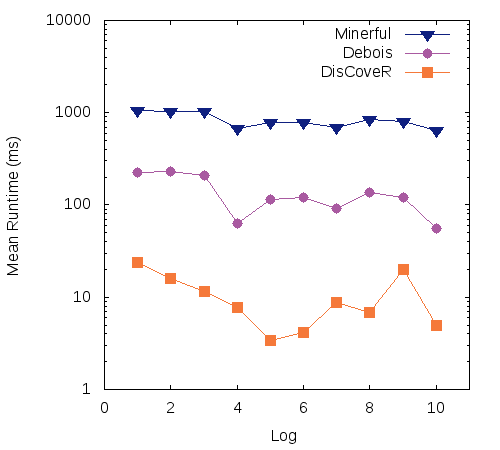}
\caption{Mean run-times in milliseconds across 100 runs on PDC 2019 training logs. \minerful~ was run with the thresholds: \emph{support} = 1.0, \emph{confidence} = 0.75, \emph{interest factor} = 0.5.}
\label{fig:run-times}
\end{figure}

\paragraph{Experimental setup} Experiments were conducted on the same set of 10 training logs from the Process Discovery Contest 2019, and were run on a Lenovo Thinkpad P50 with an Intel Xeon E3-1535M v5 2.90 GHz quad-core processor and 32G of RAM. We present mean run-times over 100 runs of mining each log. \minerful~ was parametrized with support threshold of 1.0, a confidence threshold of 0.75 and interest factor threshold of 0.5. We did \emph{not} employ the parameter tuning procedure used to achieve the results for \minerful{} in Table \ref{tab:confusion_matrices} which requires re-running the miner many times. 

The 10 logs all consist of 40 traces. Run-time results can be seen in Figure \ref{fig:run-times} as well as Table \ref{tab:run-times}, where details regarding number of activities and mean trace length are also included. 

\begin{table}[!h]
\begin{tabularx}{0.5\textwidth}{ c | r r r | c r}
&\multicolumn{3}{c |}{run-time (ms)}& &  mean\\ 
	& && & \# of & trace\\
Log & \discover~& Debois~&\minerful~& activities & length\\
\midrule
1&        23.9&     227.0&    1058.2& 45 & 17.2 \\
2&        15.8&     232.3&    1030.2& 46 & 19.0 \\
3&        11.5&     210.5&    1021.1& 48 & 12.0 \\
4&         7.7&      62.7&     678.0& 34 & 10.1 \\
5&         3.4&     114.2&     788.6& 44 & 5.3 \\
6&         4.2&     121.8&     788.8& 43 & 8.6 \\
7&         8.8&      90.8&     690.3& 35 & 12.6 \\
8&         6.8&     137.4&     837.1& 44 & 9.0 \\
9&        20.0&     121.0&     795.3& 29 & 26.4 \\
10&        4.9&      56.0&     647.8& 32 & 9.3 \\
\end{tabularx}
\caption{Mean run-times in milliseconds across 100 runs on PDC 2019 training logs, along with log statistics. \minerful~was run with the thresholds: \emph{support} = 1.0, \emph{confidence} = 0.75, \emph{interest factor} = 0.5.}
\label{tab:run-times}
\end{table}

\setlength{\tabcolsep}{3.5pt}

\begin{table*}[h!]
\begin{tabularx}{\textwidth}{ c c | c c | c c | c c | c c | c c | c c | c c | c c | c c | c c | c c | r c c }
						&					&				&\multicolumn{19}{c|}{}																																																																																		&				&				&		&\multicolumn{2}{c}{\textsc{TARGET}}	\\ 
						&					&				&\multicolumn{19}{c|}{\textbf{\textsc{Observed}}}																																																																											&				&				&		&\multicolumn{2}{c}{\textsc{TRACES}}	\\
						&					&				&\multicolumn{19}{c|}{}																																																																																		&				&				&		&Posi-	&Nega-	\\
						&					& \multicolumn{2}{c |}{$P_1$}		& \multicolumn{2}{c |}{$P_2$}		& \multicolumn{2}{c |}{$P_3$}		& \multicolumn{2}{c |}{$P_4$}		& \multicolumn{2}{c |}{$P_5$}			& \multicolumn{2}{c |}{$P_6$}		& \multicolumn{2}{c |}{$P_7$}		& \multicolumn{2}{c |}{$P_8$}		& \multicolumn{2}{ c |}{$P_9$}		& \multicolumn{2}{c |}{$P_{10}$}	&\multicolumn{2}{ c|}{Aggregate}&		&tive	&tive	\\
\midrule
\midrule
\multicolumn{2}{l |}{DisCoveR}				& \legal  		& \illegal 			& \legal		& \illegal 			& \legal  		& \illegal			& \legal  		& \illegal 			& \legal		& \illegal 				& \legal  		& \illegal 			& \legal  		& \illegal 			& \legal		& \illegal 			& \legal  		& \illegal 			& \legal		& \illegal			& \legal		& \illegal 		& MCC 	&\multicolumn{2}{c}{\textit{0.92}}\\ 
\midrule						
\textbf{\textsc{Pred-}}	& \legal			& 45			& 0 				& 45			& 0					& 45			& 0					& 47			& 6					& 45			& 3						& 45			& 0					& 44			& 8					& 43			& 2					& 45			& 3					& 44			& 8					& \textbf{448}	& \textbf{30}	&Prec.	&0.94	&0.99	\\
\textbf{\textsc{icted}}	& \illegal			& 0				& 45				& 0				& 45				& 0				& 45				& 1				& 36				& 0				& 43					& 0				& 45				& 1				& 37				& 2				& 43				& 0				& 42				& 1				& 37				& \textbf{5}	& \textbf{417}	& Recall&0.99	&0.93	\\
\multicolumn{2}{l |}{Model size}			& \multicolumn{2}{c|}{\textit{142}}	& \multicolumn{2}{c|}{\textit{189}}	& \multicolumn{2}{c|}{\textit{271}}	& \multicolumn{2}{c|}{\textit{182}}	& \multicolumn{2}{c|}{\textit{447}}		& \multicolumn{2}{c|}{\textit{412}}	& \multicolumn{2}{c|}{\textit{143}}	& \multicolumn{2}{c|}{\textit{284}}	& \multicolumn{2}{c|}{171}			& \multicolumn{2}{c|}{136}			& Acc.:  		& 96.1\% 		&$F_1$	&0.96	&0.96	\\				
\midrule
\midrule
\multicolumn{2}{l |}{Debois, et al}			& \legal  		& \illegal 			& \legal		& \illegal 			& \legal  		& \illegal			& \legal  		& \illegal 			& \legal		& \illegal 				& \legal  		& \illegal 			& \legal  		& \illegal 			& \legal		& \illegal 			& \legal  		& \illegal 			& \legal		& \illegal			& \legal		& \illegal 		& MCC 	&\multicolumn{2}{c}{\textit{0.03}}\\
\midrule						
\textbf{\textsc{Pred-}}	& \legal			& 0				& 0 				& 0				& 0					& 1				& 0 				& 0				& 0					& 12			& 7 					& 0				& 0					& 6				& 4 				& 1				& 1					& 1				& 0 				& 6				& 8					& \textbf{27}	& \textbf{20} 	&Prec.	&0.57	&0.50	\\			
\textbf{\textsc{icted}}	& \illegal			& 45			& 45 				& 45			& 45				& 45			& 44 				& 42			& 48				& 38			& 33 					& 45			& 45				& 41			& 39 				& 44			& 44				& 45			& 44 				& 37			& 39				& \textbf{427}	& \textbf{426} 	&Recall	&0.06	&0.96	\\
\multicolumn{2}{l |}{\textit{Model size}}	& \multicolumn{2}{c|}{\textit{1821}}& \multicolumn{2}{c|}{\textit{2293}}& \multicolumn{2}{c|}{\textit{2376}}& \multicolumn{2}{c|}{\textit{641}}	& \multicolumn{2}{c|}{\textit{1557}}	& \multicolumn{2}{c|}{\textit{1515}}& \multicolumn{2}{c|}{\textit{1268}}& \multicolumn{2}{c|}{\textit{1716}}& \multicolumn{2}{c|}{\textit{984}}	& \multicolumn{2}{c|}{\textit{775}}	&  Acc.: 		& 50.4\%		&$F_1$	&0.11	&0.66	\\
\midrule
\midrule
\multicolumn{2}{l |}{MINERful$^1$}			& \legal  		& \illegal 			& \legal		& \illegal 			& \legal  		& \illegal			& \legal  		& \illegal 			& \legal		& \illegal 				& \legal  		& \illegal 			& \legal  		& \illegal 			& \legal		& \illegal 			& \legal  		& \illegal 			& \legal		& \illegal			& \legal		& \illegal 		& MCC	&\multicolumn{2}{c}{\textit{0.79}}\\
\midrule						
\textbf{\textsc{Pred-}}	& \legal			& 45			& 7 				& 45			& 2					& 45			& 6 				& 48			& 5					& 45			& 21 					& 45			& 3					& 45			& 13 				& 45			& 15				& 45			& 17 				& 44			& 7					& \textbf{452}	& \textbf{96} 	&Prec.	&0.82	&0.98	\\
\textbf{\textsc{icted}}	& \illegal			& 0				& 38 				& 0				& 43				& 0 			& 39 				& 0				& 37				& 0				& 24 					& 0				& 42				& 0				& 32				& 0				& 30				& 0				& 28				& 1 			& 38				& \textbf{1}	& \textbf{351} 	&Recall	&0.98	&0.79	\\
\multicolumn{2}{l |}{\textit{Model size}}	& \multicolumn{2}{c|}{\textit{99}}	& \multicolumn{2}{c|}{\textit{99}}	& \multicolumn{2}{c|}{\textit{92}}	& \multicolumn{2}{c|}{\textit{96}}	& \multicolumn{2}{c|}{\textit{89}}		& \multicolumn{2}{c|}{\textit{99}}	& \multicolumn{2}{c|}{\textit{99}}	& \multicolumn{2}{c|}{\textit{94}}	& \multicolumn{2}{c|}{\textit{94}}	& \multicolumn{2}{c|}{\textit{97}}	&  Acc.:  		&  89.9\% 		&$F_1$	&0.90	&0.87	\\
\midrule
\multicolumn{2}{l |}{MINERful$^2$}			& \legal  		& \illegal 			& \legal		& \illegal 			& \legal  		& \illegal			& \legal  		& \illegal 			& \legal		& \illegal 				& \legal  		& \illegal			& \legal  		& \illegal 			& \legal		& \illegal 			& \legal  		& \illegal 			& \legal		& \illegal			& \legal		& \illegal 		& MCC	&\multicolumn{2}{c}{\textit{0.85}}\\
\midrule						
\textbf{\textsc{Pred-}}	& \legal			& 43			& 2 				& 45			& 1					& 44			& 4 				& 48			& 1					& 45			& 19 					& 45			& 4					& 45			& 8 				& 45			& 7					& 45			& 14 				& 42			& 7					& \textbf{447}	& \textbf{67} 	&Prec.	&0.87	&0.98	\\
\textbf{\textsc{icted}}	& \illegal			& 2				& 43 				& 0				& 44				& 1 			& 41 				& 0				& 41				& 0				& 26 					& 0				& 41				& 0				& 37				& 0				& 38				& 0				& 31				& 3 			& 38				& \textbf{6}	& \textbf{380} 	&Recall	&0.99	&0.85	\\
\multicolumn{2}{l |}{\textit{Model size}}	& \multicolumn{2}{c|}{\textit{182}}	& \multicolumn{2}{c|}{\textit{188}}	& \multicolumn{2}{c|}{\textit{186}}	& \multicolumn{2}{c|}{\textit{194}}	& \multicolumn{2}{c|}{\textit{198}}		& \multicolumn{2}{c|}{\textit{199}}	& \multicolumn{2}{c|}{\textit{199}}	& \multicolumn{2}{c|}{\textit{189}}	& \multicolumn{2}{c|}{\textit{174}}	& \multicolumn{2}{c|}{\textit{183}}	&  Acc.: 		&  91.9\% 		&$F_1$	&0.92	&0.91	\\
\midrule
\multicolumn{2}{l |}{MINERful$^3$}			& \legal  		& \illegal 			& \legal		& \illegal 			& \legal  		& \illegal			& \legal  		& \illegal 			& \legal		& \illegal 				& \legal  		& \illegal 			& \legal  		& \illegal 			& \legal		& \illegal			& \legal  		& \illegal 			& \legal		& \illegal			& \legal		& \illegal 		& MCC	&\multicolumn{2}{c}{\textit{0.84}}\\
\midrule						
\textbf{\textsc{Pred-}}	& \legal			& 42			& 1 				& 45			& 0					& 43			& 4 				& 48			& 1					& 45			& 17 					& 45			& 2					& 42			& 10 				& 44			& 10				& 45			& 8 				& 41			& 7					& \textbf{440}	& \textbf{60} 	&Prec.	&0.88	&0.97	\\
\textbf{\textsc{icted}}	& \illegal			& 3				& 44 				& 0				& 45				& 2 			& 41 				& 0				& 41				& 0				& 28 					& 0				& 43				& 3				& 35				& 1				& 35				& 0				& 37				& 4 			& 38				& \textbf{13}	& \textbf{387} 	&Recall	&0.97	&0.87	\\
\multicolumn{2}{l |}{\textit{Model size}}	& \multicolumn{2}{c|}{\textit{295}}	& \multicolumn{2}{c|}{\textit{294}}	& \multicolumn{2}{c|}{\textit{297}}	& \multicolumn{2}{c|}{\textit{296}}	& \multicolumn{2}{c|}{\textit{266}}		& \multicolumn{2}{c|}{\textit{298}}	& \multicolumn{2}{c|}{\textit{297}}	& \multicolumn{2}{c|}{\textit{288}}	& \multicolumn{2}{c|}{\textit{299}}	& \multicolumn{2}{c|}{\textit{219}}	&  Acc.:  		& 91.9 \% 		&$F_1$	&0.92	&0.91	\\
\midrule
\multicolumn{2}{l |}{MINERful$^4$}			& \legal  		& \illegal 			& \legal		& \illegal 			& \legal  		& \illegal			& \legal  		& \illegal 			& \legal		& \illegal 				& \legal  		& \illegal 			& \legal  		& \illegal 			& \legal		& \illegal 			& \legal  		& \illegal 			& \legal		& \illegal			& \legal		& \illegal 		& MCC	&\multicolumn{2}{c}{\textit{0.85}}\\
\midrule						
\textbf{\textsc{Pred-}}	& \legal			& 40			& 1 				& 45			& 1					& 43			& 4 				& 47			& 1					& 45			& 17 					& 45			& 1					& 42			& 7 				& 44			& 9					& 45			& 6 				& 41			& 7					& \textbf{437}	& \textbf{54} 	&Prec.	&0.89	&0.96	\\
\textbf{\textsc{icted}}	& \illegal			& 5				& 44 				& 0				& 44				& 2 			& 41 				& 1				& 41				& 0				& 28 					& 0				& 44				& 3				& 38				& 1				& 36				& 0				& 39				& 4 			& 38				& \textbf{16}	& \textbf{393} 	&Recall	&0.96	&0.88	\\
\multicolumn{2}{l |}{\textit{Model size}}	& \multicolumn{2}{c|}{\textit{399}}	& \multicolumn{2}{c|}{\textit{388}}	& \multicolumn{2}{c|}{\textit{391}}	& \multicolumn{2}{c|}{\textit{391}}	& \multicolumn{2}{c|}{\textit{381}}		& \multicolumn{2}{c|}{\textit{398}}	& \multicolumn{2}{c|}{\textit{359}}	& \multicolumn{2}{c|}{\textit{381}}	& \multicolumn{2}{c|}{\textit{399}}	& \multicolumn{2}{c|}{\textit{219}}	&  Acc.: 		& 92.2 \% 		&$F_1$	&0.93	&0.92	\\
\midrule
\multicolumn{2}{l |}{MINERful$^5$}			& \legal  		& \illegal 			& \legal		& \illegal 			& \legal  		& \illegal			& \legal  		& \illegal 			& \legal		& \illegal 				& \legal  		& \illegal 			& \legal  		& \illegal 			& \legal		& \illegal 			& \legal  		& \illegal 			& \legal		& \illegal			& \legal		& \illegal 		& MCC	&\multicolumn{2}{c}{\textit{0.84}}\\
\midrule						
\textbf{\textsc{Pred-}}	& \legal			& 39			& 1 				& 43			& 0					& 42			& 4 				& 47			& 1					& 45			& 28 					& 45			& 1					& 39			& 7 				& 44			& 6					& 45			& 3 				& 40			& 7					& \textbf{429}	& \textbf{47} 	&Prec.	&0.90	&0.94	\\
\textbf{\textsc{icted}}	& \illegal			& 6				& 44 				& 2				& 45				& 3 			& 41 				& 1				& 41				& 0				& 17 					& 0				& 44				& 6				& 38				& 1				& 39				& 0				& 42				& 5 			& 38				& \textbf{24}	& \textbf{400} 	&Recall	&0.95	&0.89	\\
\multicolumn{2}{l |}{\textit{Model size}}	& \multicolumn{2}{c|}{\textit{486}}	& \multicolumn{2}{c|}{\textit{476}}	& \multicolumn{2}{c|}{\textit{494}}	& \multicolumn{2}{c|}{\textit{475}}	& \multicolumn{2}{c|}{\textit{443}}		& \multicolumn{2}{c|}{\textit{499}}	& \multicolumn{2}{c|}{\textit{497}}	& \multicolumn{2}{c|}{\textit{467}}	& \multicolumn{2}{c|}{\textit{495}}	& \multicolumn{2}{c|}{\textit{435}}	&  Acc.:  		&  92.1\% 		&$F_1$	&0.92	&0.92	\\
\bottomrule
\end{tabularx}
\caption{Confusion matrices for individual data sets, each generated by separate ground truth model, in our formulation referred to as $(P_i, \mathbb{P}_{P_i})$. Precision(Prec.), Recall and $F_1$-scores are reported for which the target class is legal and illegal traces, respecively. Matthews Correlation Coefficient (MCC) is also reported. \minerful$^n$ refers to a parametrization which results in a model with fewer than $n \cdot 100$ constraints.}
\label{tab:confusion_matrices}
\end{table*}

\section{Case Study: Interactive Model Recommendation}
\label{sec:casestudies}

In this section we discuss how DisCoveR has been integrated in the \url{dcrgraphs.net} process portal as a means to provide modeling recommendations for the interactive modeling of declarative knowledge-intensive processes.
We start by briefly describing the portal and its main functionalities. We then show how process discovery has been integrated in the portal and end with a discussion on how the model recommendation functionality is used in practice.

\subsection{The DCR Process Portal}
The \url{dcrgraphs.net} process portal is a cloud-based commercial modeling solution for declarative process models, offering an extensive range of functions including process modeling, simulation, analysis, maintenance, and a wide variety of collaboration features. 
The portal has been created and is maintained by DCR Solutions, in close collaboration with researchers from the University of Copenhagen, IT University of Copenhagen and Danish Technical University.
The DCR notation, portal and DCR process engine have been applied in a range of application domains. Most notably the engine was integrated into Workzone, a case management product used by over 70\% of Danish central government institutions\footnote{\url{http://www.kmd.dk/indsigter/fleksibilitet-og-dynamisk-sagsbehandling-i-staten}} and the portal has become a cornerstone of the Ecoknow  research project\footnote{\url{https://ecoknow.org/}}, which proposes a novel digitalization strategy for Danish municipalities grounded in the declarative modeling of knowledge-intensive citizen processes. 

% Adding an asterisk makes the figure span both columns - just a suggestion
\begin{figure}[htb]
  \includegraphics[width=1\columnwidth]{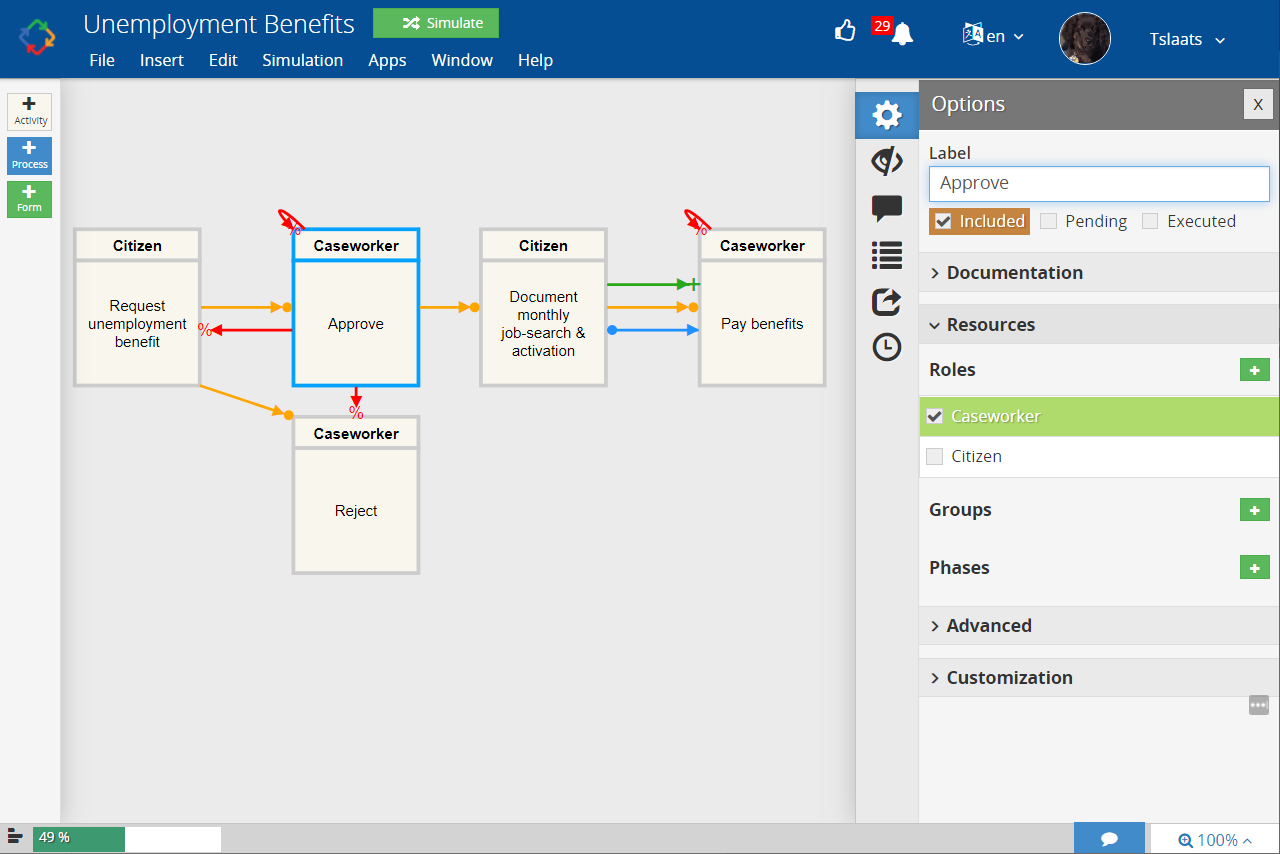}
\caption{DCR Graphs Modeling}
\label{fig:dcrportal}       % Give a unique label
\end{figure}

\begin{figure}[htb]
  \includegraphics[width=1\columnwidth]{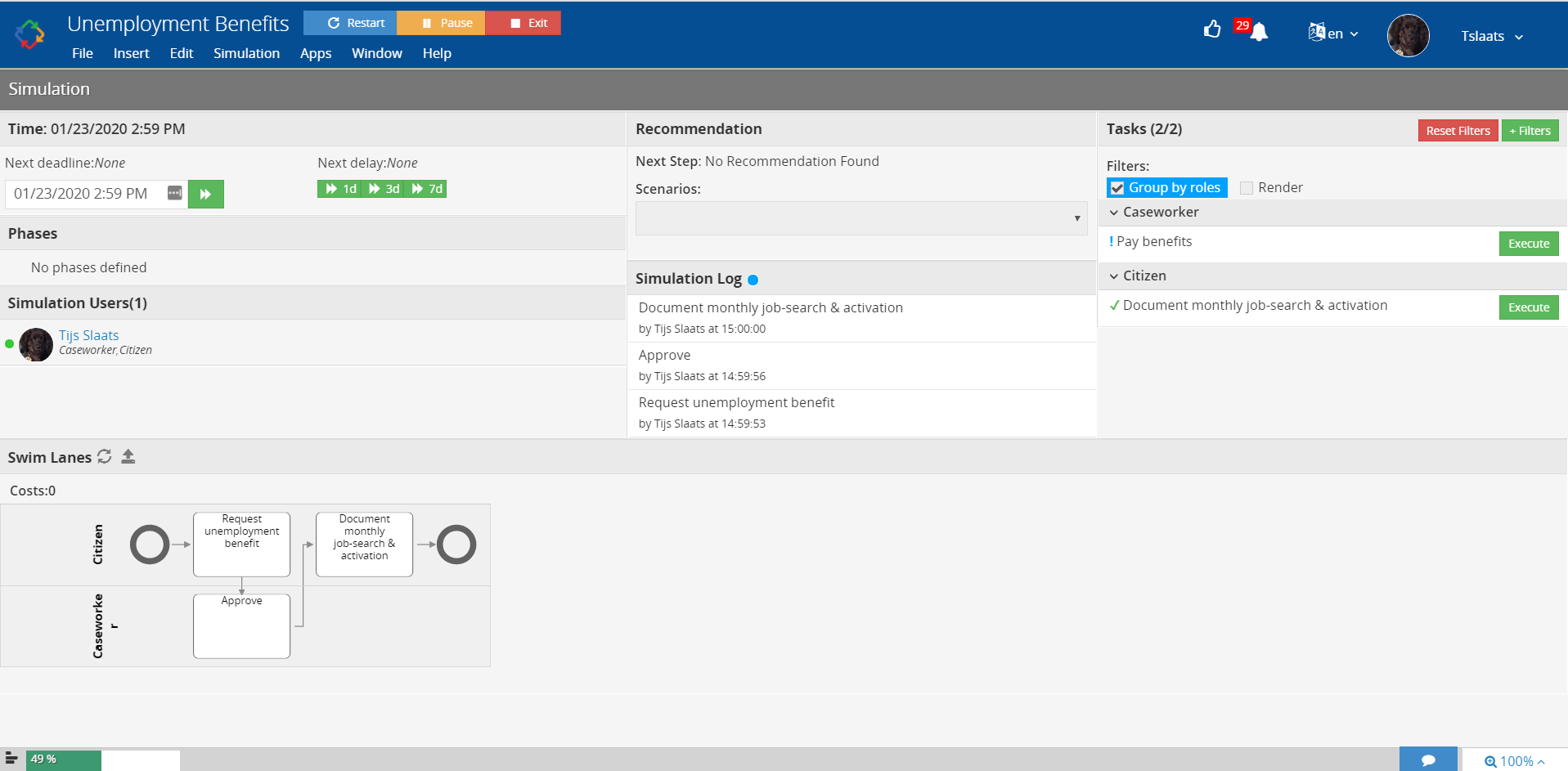}
\caption{DCR Graphs Simulation}
\label{fig:simulation}       % Give a unique label
\end{figure}

The key component of the portal is the DCR modeling tool, shown in Figure~\ref{fig:dcrportal}, which allows users to model and simulate DCR graphs. At the center of the screen is the modeling pane with the graphical representation of the DCR Graph, where activities are drawn as boxes and relations as colored arrows in a style similar to the formal syntax. Users can add and manipulate activities and relations between them directly in the modeling pane and change their details in an option panel on the right. The simulation screen is shown in Figure~\ref{fig:simulation}. The upper right of the screen shows the current task list, here the user can select which task to execute next. The middle of the screen shows recommendations for next steps and a simulation log. On the left we have a number of advanced features, such as making time steps and a list of all users involved in the simulation (collaborative simulations are supported). In the bottom of the screen the user can see a step-by-step flowchart representation of the current simulation, divided into swimlanes.

% Adding an asterisk makes the figure span both columns - just a suggestion

\subsection{Interactive process modelling through model recommendation}

\begin{figure}[htb]
  \includegraphics[width=1\columnwidth]{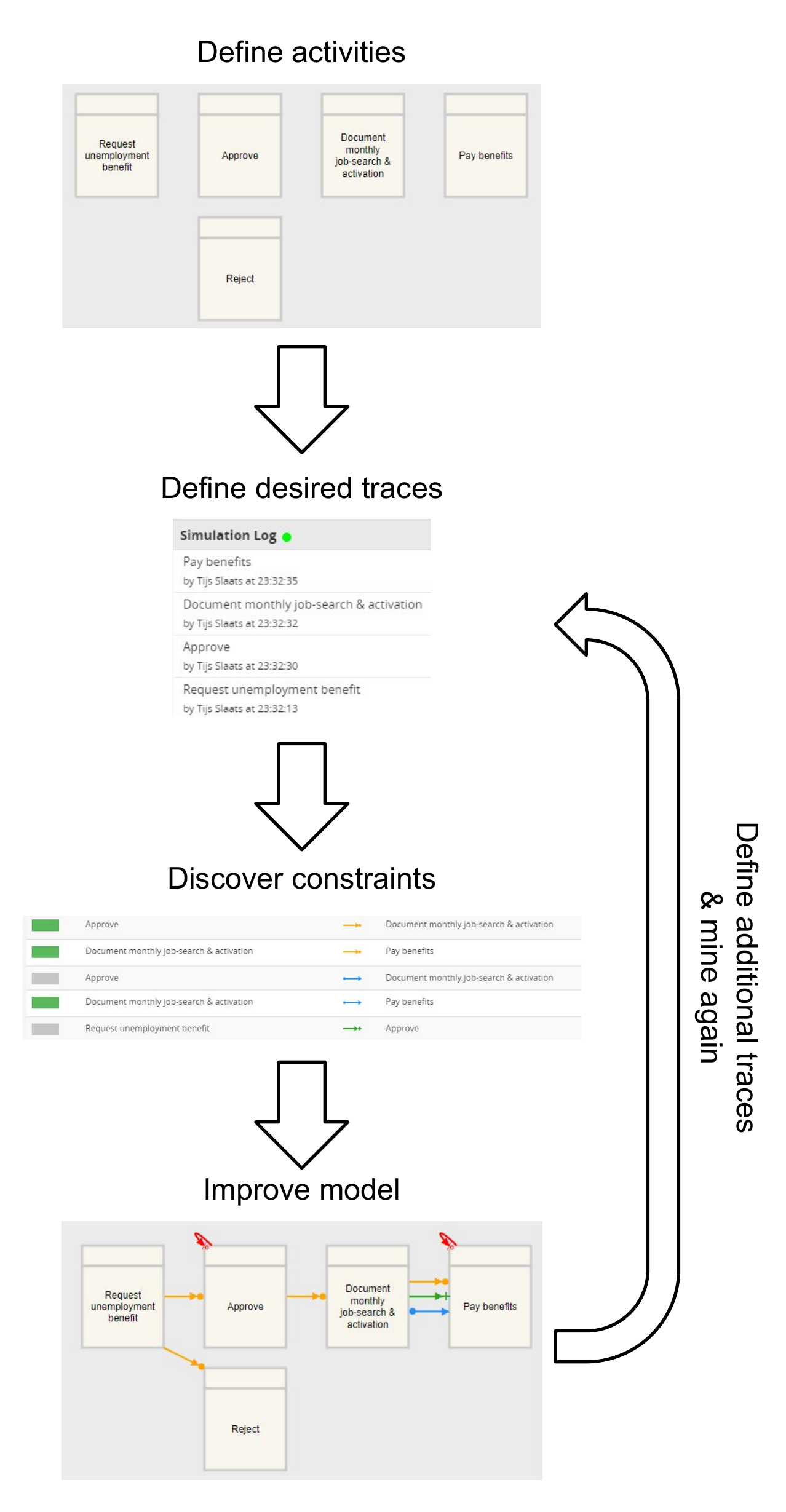}
\caption{Overview of the model recommendation approach}
\label{fig:approach}       % Give a unique label
\end{figure}

In the declarative modeling approach advocated by DCR Solutions modelers are encouraged to 1) identify the activities and roles of the process, 2) think about what common and uncommon scenarios (i.e. traces) should be supported by the process, 3) based on the scenarios determine what reasonable constraints for the process would be, and 4) ensure that the constraints do not conflict with any desired paths through the use of simulation and test-cases~\cite{10.1007/978-3-319-98648-7_3}.
The identification of constraints in step 3 has been identified as the most challenging for users because it requires a firm grasp of the semantics of DCR Graphs. While test cases and simulation can be used to retroactively check that no conflicting constraints have been introduced, they are not helpful for identifying suitable constraints directly. As a result, novice users often use a fairly inefficient trial-and-error approach where they try a constraint, check how it behaves under simulation and then update their model accordingly.

We introduced process discovery as an alternative to this trial-and-error approach. In this new setting, the portal supports the user by having an algorithm automatically propose suitable relations based either on an existing event log, and/or the traces that were identified during step 2 of the previously sketched modeling method.

Figure~\ref{fig:approach} provides an overview of the adapted approach: we start by by identifying the activities of the process and modeling these directly in the portal. In the next step we run simulations on these activities (recall that following the declarative paradigm, these simulations are entirely unconstrained and any trace can be generated). We store the traces generated during the simulation and use these as input for the following step, where we use DisCoveR to identify constraints based on the generated traces. Finally the user can improve on their model and potentially run more simulations which can be used for additional process discovery, possibly finding additional constraints that were not found for the initial traces.

% Adding an asterisk makes the figure span both columns - just a suggestion
\begin{figure}[htb]
  \includegraphics[width=1\columnwidth]{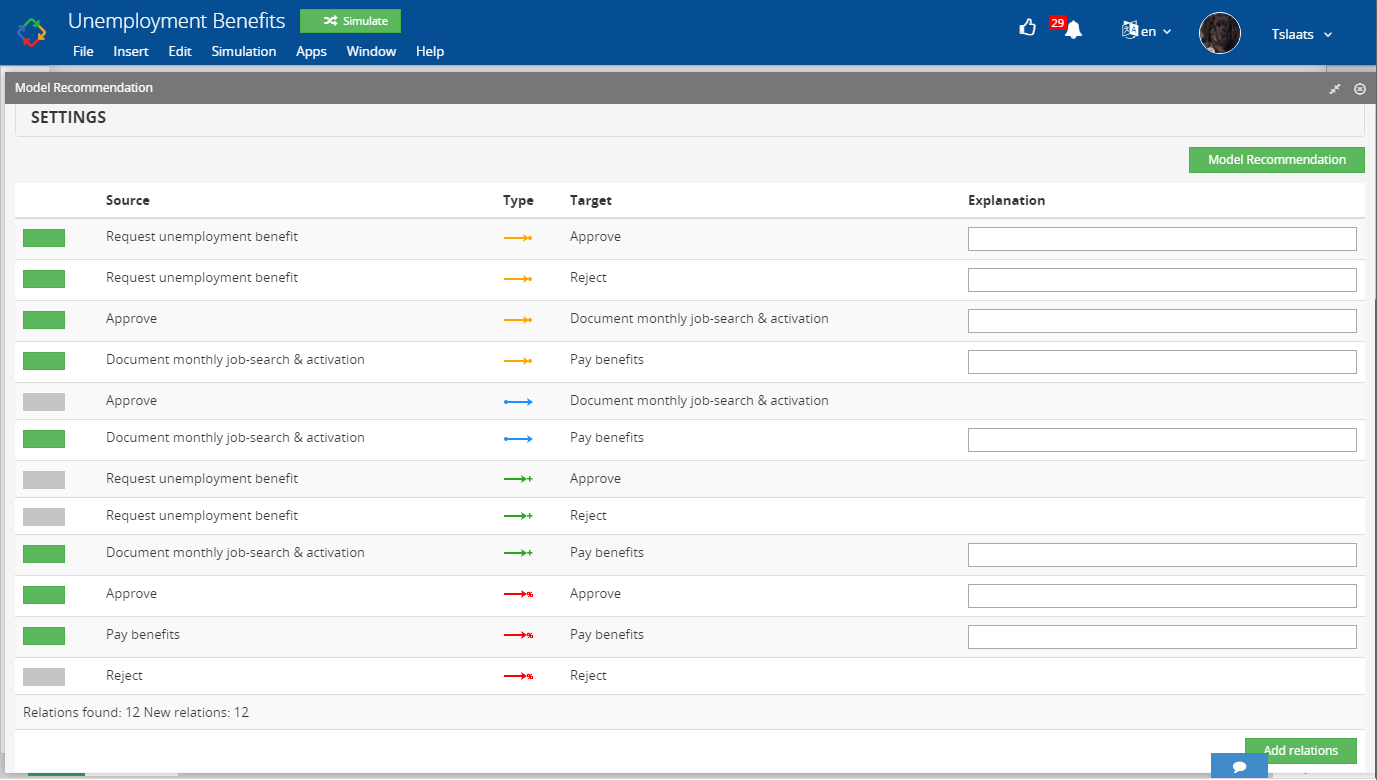}
\caption{Model Recommendation}
\label{fig:modelrecommendation}       % Give a unique label
\end{figure}

The model recommendation screen is shown in Figure~\ref{fig:modelrecommendation} and fairly straightforward: the user is shown which relations were found between which activities and can select those they wish to add through the box on the left. The user can also enter an explanation for the relation (i.e. why was it added or left out), this enables rationale management of the model and allows other users to follow the modeler's reasoning. In addition, we plan to use this information in the future to improve upon the discovery algorithm. By clicking \emph{Add Relations}, all selected relations are added to the model.

\subsection{Discussion}

Since the integration of DisCoveR into the DCR Graphs portal, DCR Solutions has been actively conducting workshops with users where the new methodology is demonstrated and used.
The inclusion of process mining in the modeling task was embraced enthusiastically by users and has been (informally) observed to lower the complexity of the modeling task.

In the traditional modeling exercise, users that are more familiar with BPMN and/or flow charts are often hampered by the novelty of the notation, e.g. they will be unclear on what the different relations mean and how to use them. In particular, the fact that arrows do not indicate flow, but logical relations between the activities can lead to confusion.
Using model recommendations, on the other hand, has allowed DCR Solutions to ask the users questions based on the recommended relations such as, “Is it true that approval is a condition for providing documentation?” or, “Is it true approval removes the ability to reject?”.

In essence, model recommendation has managed to bridge an important gap between the consultant and user: in the past, the users were new to the notation, the consultants to the process. This made building a common understanding about the process a time intensive task. Model recommendation closes this divide by, on the one hand, helping the consultant better understand the process and, on the other, providing the user with examples of the notation that are uniquely fitting to their own domain.

%Such questions seems easier for users to relate to rather than adding an orange or blue relation between activities in the diagram. The user is also guided to activities where the relations probably need to be added.
%
%Instead of having to explain how to add relations and what they mean, we show a list of possible relations to people (se Per Andreasens – Globeteam) and ask them if it correct that X-Y – and adds that as an explanation for the relation.
%<image – ask + explanation>
%Notice – we’ve made providing an explanation mandatory in the UI to ensure people think about the relations added.
 
%Even if we have data in the graph, the model recommendation finds reasonable relations.
 
The high accuracy of the algorithm has also been noted in practice: even for processes that include other perspectives than just control-flow (e.g. decisions depending on contextual data), the algorithm has been noted to be highly successful in recommending relevant relations that improved the users' understanding of the process.

The integration of the algorithm in the commercial tools was relatively effortless: the front-end of the model recommendation was developed rapidly at DCR Solutions through existing plugin support for the portal. The algorithm itself was simply deployed as a cloud service by the researchers. Because of a long history of close collaboration between the two parties, the details of the interface between these two components and a general understanding of how the system should work was fleshed out quickly over two meetings and a few emails.
 
%We are planning to collect recommended relations and replies to use machine learning to improve accuracy of the recommendations.

%The DCR Process Methodology seems to work quite well when we need to identify processes with a group of people. Identifying roles and activities, i.e. what work we do and who is involved, rarely raises any discussions among the group. Then asking each individual to add 2-3 scenarios they have experienced, maybe included forbidden scenarios that they might have experienced but would like to avoid in the future, is also something most people grasp quite fast and again rarely raise any discussions.

%\input{discussion}

\section{Conclusion}
\label{sec:conclusion}

In this paper we presented DisCoveR, a declarative miner for DCR Graphs based on the ParNek algorithm. 
We formally defined the underlying algorithm and how it has been implemented using an acute mapping to bit vector operations, yielding a highly efficient process discovery tool.
We evaluated the miner using a traditional classification task and computed the standard machine learning measures of accuracy (96.1\%), precision (0.94 on positive traces, 0.99 on negative traces), recall (0.99 on positive traces, 0.93 on negative traces), F1 (0.96 on each) and MCC (0.92). We show that DisCoveR out competes all other declarative miners under consideration on each of these measures. In addition an analysis of its run-time shows that it is one order of magnitude faster than the state-of-the-art in DCR Graphs discovery and two orders of magnitude faster than the state-of-the-art in Declare discovery.
Finally, we showed how the tool has been integrated in a commercial modeling tool and discuss how its integration has significantly improved the modeling experiences of its users. 

\subsection{Future Work}

Several avenues exist for future work in mining DCR Graphs from event logs. So far, we have considered only the control flow of processes. Incorporating timing, data, and resource perspectives is extermely relevant for many real-world scenarios and one of the primary requests made by DCR Solutions. 

Also, we restricted our hypothesis space to graphs with the same simple initial marking in which all events are enabled. This is due to the complicated interactions arising with other relations when excluding a source event. Considering different initial markings would enable the discovery of more complex models, but also enlarge the hypothesis space and increase the danger of overfitting. 

In order to control more explicitly for overfitting and quantify the tradeoff between inductive bias and complexity, a formulation of regularization functions for classes of DCR Graphs is an important next step. This is not entirely straightforward due to the non-monotonic nature of DCR Graphs \cite{debois2015safety}, rendering simple relation counting more or less meaningless for regularization purposes. 

As described in the case study, users of the \url{dcrgraphs.net} portal are not only able to define positive scenarios, but also undesired scenarios. The use of negative input data in process discovery has so far been mostly ignored based on the assumption that such data is not available. Having negative scenarios provided by the portal offers a unique opportunity to develop new algorithms that take negative examples as input and thereby produce more relevant models. We observe that DisCoveR has a noticeably lower recall on negative than positive traces and hypothesize that the ability to analyze negative examples of traces will help us improve on this aspect of the accuracy of the tool.

Finally, there remain certain points in the ParNek algorithm in which choices are currently taken in a naive manner (e.g.~$ChooseOneRelation$). This decision point should be framed as a proper optimization problem. In fact, framing DCR Graph mining properly as an optimization task would open a powerful set of tools from the general optimization literature.

\label{sec:futurework}

%\begin{acknowledgements}
%\end{acknowledgements}


%%% NOTE %%%
% The guidelines for the International Journal on Software Tools for Technology Transfer do not specify a specific bibliographic style or whether citations should be consecutive (in order of occurrence) or alphabetical, and published articles seem to follow both.
%%%%%%%%%%%%

\bibliographystyle{spmpsci}       % mathematics and physical sciences (ALPHABETICAL SORT)
\bibliography{biblio,bib1,bib2,bib3,bib4}   % name your BibTeX data base

\end{document}